\documentclass{article}
\usepackage[numbers,sort&compress]{natbib}
\usepackage[preprint]{neurips_2026}

\usepackage[utf8]{inputenc} 
\usepackage[T1]{fontenc}    
\usepackage{hyperref}       
\usepackage{url}            
\usepackage{booktabs}       
\usepackage{amsfonts}       
\usepackage{nicefrac}       
\usepackage{microtype}      
\usepackage[table]{xcolor}  
\usepackage{graphicx}
\usepackage{algorithm}
\usepackage{algorithmic}

\title{Look Before You Reason: Implicit Visual \\ Thinking for Efficient Multimodal Reasoning}

\author{%
  \normalfont Zerui Chen, Changrui Chen, Fei Ni, and Jiankang Deng \\[4pt]
  MVP Lab, Imperial College London \\[2pt]
  \url{https://zerchen.github.io/projects/lore.html}
}

\begin{document}

\maketitle

\begin{abstract}
Vision-language models (VLMs) have advanced rapidly in multimodal reasoning, yet they remain limited on tasks requiring fine-grained understanding of small or easily overlooked image regions. Recent \emph{thinking with images} approaches mitigate this limitation by allowing models to inspect images through operations such as zooming, cropping, or code execution. While effective, their multi-turn interactions with VLMs introduce redundant token computation and slow inference, and separate region inspection from the reasoning step that depends on it. To address these issues, we propose \textbf{Lo}ok Before You \textbf{Re}ason (LoRe), an implicit visual thinking framework for efficient visual reasoning within a single-turn VLM interaction. Given an image and a question, LoRe predicts the most relevant image region, aligns it with the corresponding image patch features, and strengthens the associated key-value (KV) cache entries, enabling subsequent decoding to directly exploit the selected visual cues without additional visual inputs or explicit tool calls. We further employ reinforcement learning to guide the VLM toward identifying informative image regions and leveraging them during reasoning. To support fine-grained evaluation of multimodal reasoning, we introduce COCO-ObjQA, a large-scale benchmark for object-centric multimodal reasoning. Experiments on three challenging multimodal benchmarks demonstrate that LoRe consistently improves VLM reasoning performance with low additional inference overhead.
\end{abstract}

\section{Introduction}
\label{sec:intr}
Recent vision-language models (VLMs) have made rapid progress in multimodal reasoning, yet they continue to struggle when answers depend on small, localized, or easily overlooked image regions. To address this limitation, the emerging \emph{thinking with images} paradigm~\cite{su2025thinking,zheng2025deepeyes,li2025dyfo,su2025openthinkimg,wu2025vtool,zhang2025adaptive,wang2025visuothink} treats visual input not as static context, but as evidence to be actively processed through operations such as zooming, cropping, or the use of other visual tools. By leveraging multimodal chain-of-thought (CoT)~\cite{wei2022chain} elicited through supervised fine-tuning (SFT) or reinforcement learning (RL)~\cite{guo2025deepseek,sutton1998reinforcement}, these approaches have demonstrated promising performance on diverse multimodal reasoning tasks.

Despite their improved reasoning capabilities, existing \emph{thinking with images} approaches remain far from efficient. At certain interaction steps with the VLM, these approaches explicitly issue a tool call to request a visual operation such as zooming or cropping, then wait for the operation to complete and receive the processed visual observation before resuming reasoning. Such back-and-forth between the VLM and external visual tools repeats across multiple turns until a final answer is reached.

This paradigm suffers from two fundamental limitations. First, it is computationally expensive: intermediate visual observations and reasoning chains must be tokenized and processed by the VLM in subsequent turns, leading to substantial token consumption and high inference latency. Second, it introduces a temporal mismatch between visual inspection and reasoning. When the model decides to examine a specific image region, the resulting visual evidence only becomes available in the next interaction turn, rather than in the current reasoning step that prompted the inspection. Consequently, critical visual evidence is incorporated too late to support efficient multimodal reasoning.

\begin{figure}[t]
  \centering  
  \includegraphics[width=0.91\linewidth]{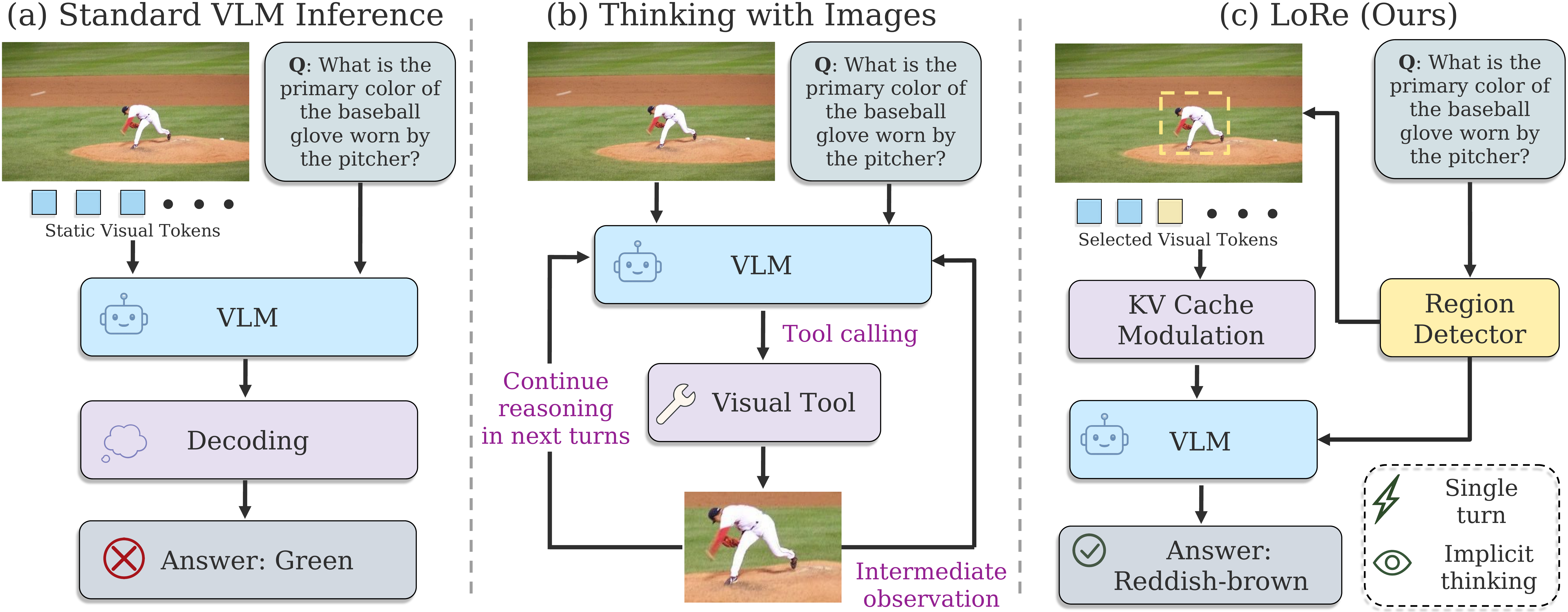}
  \vspace{-0.15cm}
    \caption{\textbf{Comparison of visual reasoning paradigms.} (a) Conventional VLMs rely on static visual tokens, which limits reasoning ability, particularly for small regions. (b) \emph{thinking with images} approaches depend on costly multi-turn visual manipulation, incurring significant inference overhead. (c) LoRe achieves efficient single-turn implicit visual reasoning via lightweight KV cache modulation.}

  \vspace{-0.5cm}
  \label{fig:overview}
\end{figure}

To alleviate these limitations, as shown in Figure~\ref{fig:overview}(c), we introduce \textbf{Lo}ok Before You \textbf{Re}ason (LoRe), a novel implicit thinking algorithm for efficient reasoning requiring only a single interaction turn with VLMs. Given an image and question, LoRe first identifies the most relevant region and strengthens its visual evidence during subsequent token decoding. Unlike prior methods relying on external tools and multiple interaction turns to generate intermediate visual inputs, LoRe leverages VLMs' inherent capabilities to unify visual manipulation and reasoning within a single inference process.

Specifically, LoRe performs implicit thinking through lightweight key-value (KV) cache modulation. After identifying the most relevant region, LoRe maps it to corresponding image patch features and amplifies their entries in the KV cache. It allows subsequent decoding to focus more on the selected visual evidence. In this way, crucial visual evidence can directly influence the current reasoning step, without modifying the input image or introducing additional visual tokens. We further train VLMs with RL to encourage the model to identify the region that supports the final answer, enabling it to learn when and where to look while immediately leveraging the discovered evidence for reasoning.

To comprehensively evaluate LoRe's reasoning ability, we construct COCO-ObjQA, an object-centric benchmark derived from object-level annotations in the COCO dataset~\cite{lin2014microsoft}. Existing benchmarks such as V*~\cite{wu2024vstar} and HR-Bench~\cite{wang2025divide} remain limited in scale and coverage. In contrast, COCO-ObjQA contains more than 8,000 evaluation samples, enabling controlled evaluation of a model's ability to reason over visual targets of varying sizes. It complements existing benchmarks and more reliably exposes failure cases, particularly for small objects. Extensive experiments demonstrate that LoRe consistently improves VLM reasoning across three challenging benchmarks with low inference overhead. Our contributions are summarized as follows:

\noindent$\bullet$ We propose LoRe, an implicit thinking framework that unifies visual manipulation and reasoning within a single inference turn. By amplifying informative image tokens through KV cache modulation, LoRe enables efficient reasoning without explicit tool calls or additional visual tokens.

\noindent$\bullet$ We employ reinforcement learning to train VLMs to identify the most task-relevant image region and immediately leverage the discovered visual evidence within the same reasoning turn, enabling grounded and efficient multimodal reasoning without temporal mismatch.

\noindent$\bullet$ We curate COCO-ObjQA, a large-scale benchmark to enable controlled evaluation of fine-grained visual reasoning. Extensive experiments demonstrate that LoRe consistently improves VLM reasoning abilities across three challenging multimodal benchmarks with low inference overhead.

\section{Related Work}
\label{sec:rel}
\textbf{Large Vision-language Models.} Recent years have witnessed remarkable progress in large vision-language models~\cite{li2024llava,liu2023visual,bai2025qwen3,wang2025internvl3} (VLMs). State-of-the-art VLMs typically consist of a vision encoder~\cite{dosovitskiy2021image,yinxie_2025_rice,fini2025multimodal,tschannen2025siglip} and a large language model~\cite{liu2024deepseek,Touvronetal2023,bai2023qwentechnicalreport} (LLM). Early VLMs, such as Flamingo~\cite{Alayracetal2022} and BLIP~\cite{Lietal2022,Lietal2023}, adopt the classic encoder-decoder architecture~\cite{vaswani2017attention}, where visual features are first encoded and textual representations cross-attend to them at each token-generation step, leading to relatively slow inference. By contrast, recent VLMs~\cite{liu2023visual,li2024llava,bai2025qwen3,wang2025internvl3} typically employ a simpler decoder-only architecture: visual features are projected into the language embedding space and processed jointly with text tokens via self-attention for autoregressive generation. This architectural paradigm is compatible with highly optimized LLM inference engines~\cite{kwon2023efficient,zheng2024sglang}, thereby substantially improving inference efficiency. Despite impressive progress, existing VLMs still exhibit limited reasoning ability~\cite{chen2024measuring,huynh2025vision}, as they largely treat visual content as static context for the language model~\cite{wang2025pixel}. As a result, the model reasons over compressed visual tokens without actively revisiting or emphasizing task relevant regions during inference.

\textbf{Reasoning in Vision-language Models.} To improve the reasoning ability of VLMs, recent studies have explored multimodal chain-of-thought reasoning~\cite{wei2022chain}, often elicited through supervised fine-tuning or reinforcement learning~\cite{guo2025deepseek,sutton1998reinforcement}. Earlier efforts mainly focus on eliciting step-by-step textual rationales or decomposing complex visual questions into intermediate reasoning steps, thereby encouraging VLMs to perform more deliberate inference over visual inputs~\cite{zhang2024multimodal,wang2024mementos}. Orthogonal to prompting-based reasoning, another line of work improves visual reasoning by enhancing fine-grained perception, for example through high-resolution encoding, adaptive image partitioning, or region-level visual grounding~\cite{wu2024vstar,wang2025divide,luo2024feast}. Inspired by OpenAI o3~\cite{openai_o3_o4mini_system_card_2025}, the emerging \emph{thinking with images} paradigm treats visual inputs as dynamic evidence to be actively inspected during reasoning. Existing methods typically rely on external tools~\cite{zheng2025deepeyes,wu2025reinforce,su2025openthinkimg}, executable programs~\cite{zhang2025thyme,suris2023vipergpt,gupta2023visual}, or generative imagination~\cite{chern2025thinkinggeneratedimages,li2025imagine,zhang2025latentsketchpad} to manipulate visual information. These methods share the intuition that visual reasoning should not be performed solely over a fixed set of compressed image tokens, but should instead allow the model to seek additional visual evidence when needed. While effective, these explicit \emph{thinking with images} approaches usually require multi-turn interactions, where intermediate visual observations and reasoning traces are repeatedly generated, tokenized, and processed by VLMs. This leads to substantial token overhead, high inference latency, and a temporal mismatch between visual inspection and reasoning. LoRe instead predicts a region and amplifies its existing value-cache entries within the same generation request. This targets evidence selection within an encoded image rather than the acquisition of new visual observations.

\textbf{Attention and representation steering.} PASTA~\cite{zhang2024pasta} reweights selected LLM attention heads toward user-specified text. ClearSight~\cite{yin2025clearsight} enhances visual signals during middle-layer modality fusion to reduce object hallucination, while ICT~\cite{chen2025ict} applies image- and object-level activation shifts to selected head outputs. LoRe shares the goal of strengthening relevant evidence but predicts a question-conditioned spatial region online and updates its live value-cache entries. It does not directly reweight attention probabilities or require user-marked regions. Decoding resumes in the same request with native optimized attention kernels.

\section{Methodology}

\begin{figure}[t]
    \centering
    \includegraphics[width=0.95\linewidth]{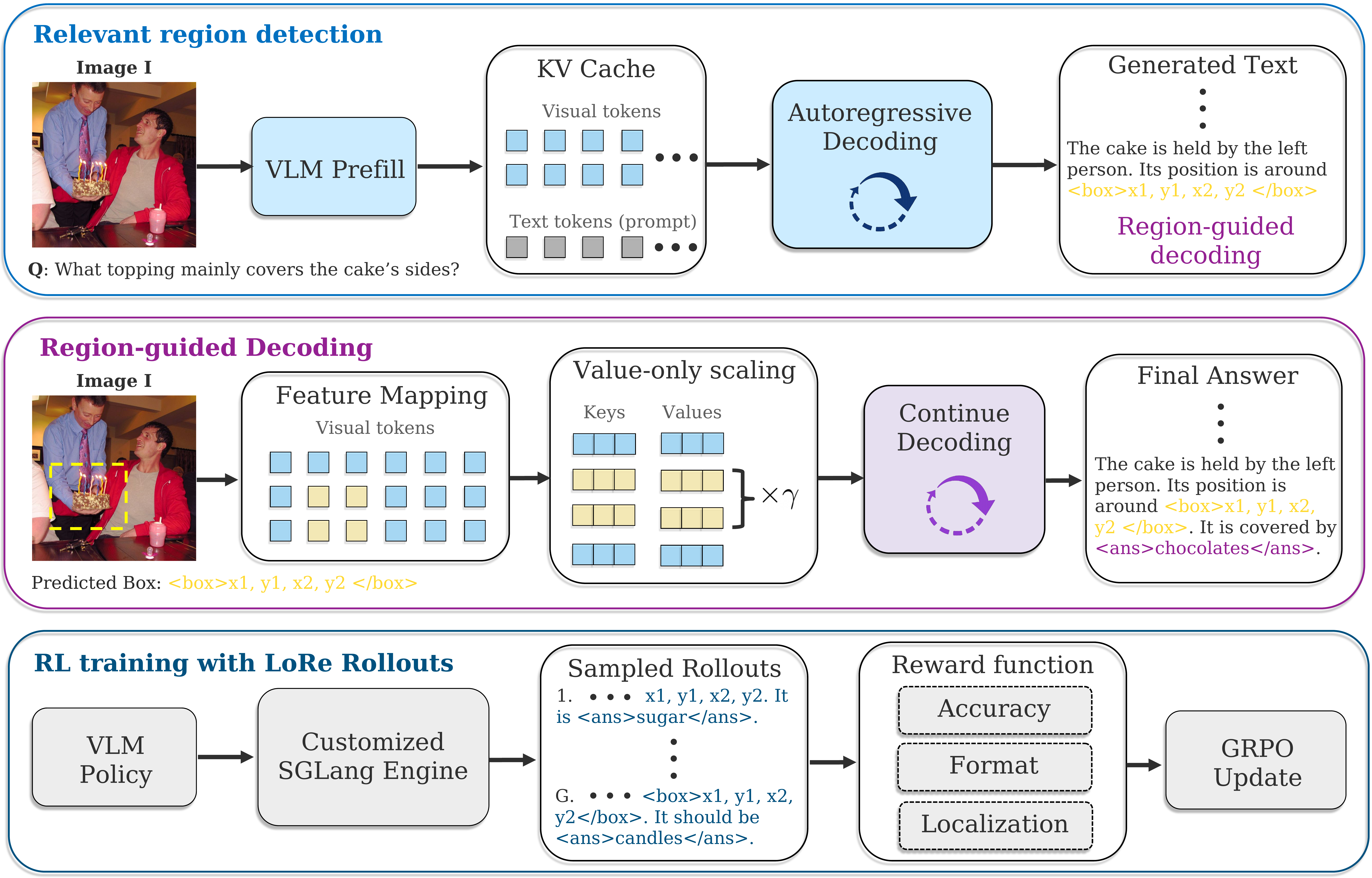}
    \vspace{-0.2cm}
    \caption{\textbf{Overview of LoRe.} The VLM emits a region as coordinate tokens. The region detector denotes this generation-and-parsing process, not a separate detection network. LoRe maps the box to visual-token positions and scales only their value-cache entries before decoding continues in the same request. Keys, position IDs, and token order remain unchanged. No crop, additional visual tokens, or second VLM call is required. RL optimizes these trajectories with answer, format, and localization rewards.}
    \vspace{-0.5cm}
    \label{fig:lore_framework}
\end{figure}

\label{sec:method}
LoRe combines question-conditioned region selection with value-cache modulation within a single generation request. Section~\ref{sec:implicit} describes this mechanism, Section~\ref{sec:training} introduces its RL training procedure, and Section~\ref{sec:data} presents COCO-ObjQA for evaluation across target-object sizes.

\subsection{Implicit visual thinking}
\label{sec:implicit}

\textbf{Preliminaries for VLM decoding.} Figure~\ref{fig:lore_framework} gives an overview of LoRe. The VLM prediction
process generally consists of two phases: prefilling and autoregressive generation. During
prefilling, the VLM tokenizes the multimodal prompt into a single sequence of
visual and text tokens, and each transformer layer stores the key-value
representations of all prompt tokens in its KV cache. During autoregressive
generation, the model forwards only the newly generated token at each step,
attends to the cached prefix, and appends the new token's key-value
representations to the cache. For a prompt of length $m$, the layer-$\ell$
cache after prefilling and after generating $t$ tokens can be written as:
\[
    C_\ell^{(0)}=\{(k_{\ell,i},v_{\ell,i})\}_{i=1}^{m}, \qquad
    C_\ell^{(t)}=C_\ell^{(t-1)}\cup\{(k_{\ell,m+t},v_{\ell,m+t})\}.
\]
This conventional mechanism makes decoding efficient, but it also fixes the visual evidence
after prefilling. Subsequent tokens can attend to the static visual features, and the
model has no native operations for amplifying the influence of a region that
becomes salient during reasoning.

\textbf{Relevant region prediction.} Given an image $I$ and a question $Q$, we append an instruction asking the VLM to first localize the relevant region and then answer. The VLM itself generates ordinary text tokens in the format \texttt{<box>}x1, y1, x2, y2\texttt{</box>}, followed by reasoning and an \texttt{<ans>}...\texttt{</ans>} answer span. There is no separate detection head or external detector. Qwen2.5-VL uses image-pixel coordinates, while Qwen3-VL uses its native $[0,1000]$ coordinate convention. Thus, the region control signal is explicit, whereas visual focusing is performed internally on the cache.

After prefilling, a lightweight processor checks the generated sequence every four tokens, searching the latest 128 tokens for a complete box span. It parses the coordinates as $R_q=(x_1,y_1,x_2,y_2)$, converts them to image coordinates, and clamps them to the image boundary. The default configuration applies one cache update when a usable region is found, then resumes the same request.

If the span is missing or malformed, or the clamped box cannot be mapped to visual tokens, the cache remains unchanged and standard decoding continues without retrying or discarding the sample. All reported accuracies include these fallback cases. A valid but inaccurate box still triggers modulation and can amplify irrelevant evidence.

\textbf{Region-guided decoding.} Once the relevant region $R_q$ is detected, LoRe uses it to guide the remaining decoding process without altering the image input or starting a new VLM call. The key step is to convert the predicted box into the visual-token positions that have already been stored in the KV cache during prefilling. Let $\mathcal{G}=(H,W)$ be the effective visual-token grid after patch merging, $(W_I,H_I)$ be the image size, and $p_i$ be the sequence position of the $i$-th visual token. For $R_q=(x_1,y_1,x_2,y_2)$, LoRe first converts normalized coordinates to the image coordinate system if needed, clamps the box to the image boundary, and selects the covered grid cells:
\[
    \mathcal{P}(R_q)=\{p_{vW+u}\mid u\in U_q,\;v\in V_q\}.
\]
Here $U_q=[\lfloor x_1W/W_I\rfloor,\lceil x_2W/W_I\rceil)$ and $V_q=[\lfloor y_1H/H_I\rfloor,\lceil y_2H/H_I\rceil)$ denote the horizontal and vertical grid ranges, with zero-based grid indices. We read \texttt{image\_grid\_thw} after dynamic-resolution preprocessing, account for spatial patch merging, and verify that the effective grid matches the visual-token count before mapping cells to row-major sequence positions. The mapping therefore adapts to image size and aspect ratio rather than assuming a fixed grid. Very small regions remain subject to patch quantization.

Given $\mathcal{P}(R_q)$, LoRe scales the selected value-cache entries once in every decoder layer $\ell\in\mathcal{L}$, leaving the keys, position IDs, token order, and attention masks unchanged:
\[
    v'_{\ell,j} =
    \left\{
    \begin{array}{ll}
    \gamma v_{\ell,j}, & j\in\mathcal{P}(R_q),\\
    v_{\ell,j}, & j\notin\mathcal{P}(R_q),
    \end{array}
    \right.
    \qquad \gamma\geq1.
\]
$\gamma$ is the amplification factor, with $\gamma=1$ disabling modulation. For a fixed query $q$, let $\alpha_j=\mathrm{softmax}_j(q^\top k_j/\sqrt{d})$ and $o=\sum_j\alpha_jv_j$, omitting layer and head indices. The intervention gives
\[
    o'=o+(\gamma-1)\sum_{j\in\mathcal{P}(R_q)}\alpha_jv_j.
\]
Thus, value scaling amplifies attended evidence without directly changing $\alpha_j$ or rotary-encoded keys. Later attention can change indirectly as the modified residual state affects subsequent queries. In the pre-normalized decoder, subsequent normalization constrains magnitude but does not undo changes in residual direction or composition. All unselected tokens remain available, although excessive scaling can overwhelm global context. This is evidence amplification, not a recovery of details lost during vision encoding. Tokens with negligible query--key affinity remain weak contributors.

Value scaling is a linear intervention compatible with native optimized attention kernels. Scaling keys would also change logits in a query-dependent manner, increasing positive logits while making negative ones more negative. An additive logit bias offers direct attention reweighting but requires a different attention implementation. LoRe uses value-only scaling by default to preserve the existing attention computation and keep overhead low.

\subsection{Reinforcement learning for LoRe}
\label{sec:training}

\textbf{Policy rollouts with LoRe.} The goal of RL training is to make the model learn a coupled behavior: it should predict a region that supports the question, and it should benefit from that region when generating the final answer. Therefore, rollouts must use the same inference procedure described in Section~\ref{sec:implicit}, rather than treating localization as an auxiliary output that is detached from decoding. Each training sample contains an image-question pair $x=(I,Q)$, a ground-truth answer $a$, and metadata such as answer type $\tau$ and problem type $z$. The policy first produces a region span and then continues generation under LoRe's region-guided cache modulation, so the final response is evaluated after the model has actually used the detected visual evidence.

To implement this rollout process efficiently, we extend the SGLang engine~\cite{zheng2024sglang} with a dynamic value-boost processor. During generation, the processor monitors the token stream, parses completed box spans, maps the detected region to visual-token positions, applies the cache modulation inside the inference backend, and then resumes decoding in the same request. For each prompt, we sample $G$ trajectories from the VLM policy $\pi_{\theta_{\mathrm{old}}}$, which provides the training data for RL.

\textbf{Reward function.} Following recent visual RL studies that rely on verifiable rewards~\cite{wu2025reinforce}, we use a rule-based reward that prioritizes final-task correctness while providing weak supervision for relevant region detection. The main term $r_{\mathrm{acc}}(y,a,\tau)$ evaluates the final answer according to the answer type $\tau$: multiple-choice answers are judged by the final answer, while numerical and OCR answers use their own matching rules. This term provides the dominant impact on LoRe's reasoning ability.

We add two auxiliary terms to stabilize the LoRe trajectory. The format reward $r_{\mathrm{fmt}}$ checks for exactly one non-empty \texttt{<ans>}...\texttt{</ans>} span and balanced optional reasoning delimiters. Region supervision comes from $r_{\mathrm{loc}}$: it gives a box-presence bonus for visually grounded samples and adds parseability and center-proximity terms for bbox-supervised samples. Thus, the VLM learns coordinate generation through these rewards rather than a separate detector loss. The final reward is:
\[
    R=\min\{1,\;r_{\mathrm{acc}}+(1-r_{\mathrm{acc}})(0.03\,r_{\mathrm{fmt}}+r_{\mathrm{loc}})\}.
\]
This residual form prioritizes correctness: correct answers are capped at one
while localization bonuses remain bounded, preventing them from dominating the
answer reward.

\textbf{GRPO update.} We adopt Group Relative Policy Optimization (GRPO)~\cite{guo2025deepseek}, following recent reinforcement learning approaches for \emph{thinking with images}~\cite{su2025openthinkimg,zheng2025deepeyes,wu2025vtool,wu2025reinforce}. For each prompt $x$, we sample a group of LoRe trajectories $\{y_i\}_{i=1}^{G}$ from $\pi_{\theta_{\mathrm{old}}}$ and compute rewards $\{R_i\}_{i=1}^{G}$. The reward of each trajectory is converted into a clipped score $S_i=\mathrm{clip}(\alpha(R_i+b),-c,c)$ and then normalized within the same prompt group to obtain $A_i=(S_i-\mu_x)/(\sigma_x+\epsilon)$. This relative advantage measures whether a trajectory finds and uses visual evidence better than other candidates for the same question.

With $\rho_{i,t}=\pi_\theta(y_{i,t}\mid x,y_{i,<t})/\pi_{\theta_{\mathrm{old}}}(y_{i,t}\mid x,y_{i,<t})$ and $\bar\rho_{i,t}=\mathrm{clip}(\rho_{i,t},1-\epsilon_{\ell},1+\epsilon_h)$, we minimize:
\[
    \mathcal{L}_{\mathrm{GRPO}}=-\mathbb{E}\!\left[\frac{1}{G}\sum_{i,t}T_i^{-1}
    \min(\rho_{i,t}A_i,\bar\rho_{i,t}A_i)\right]+\beta D_{\mathrm{KL}}.
\]
Here the expectation is over prompts and rollout groups, $T_i$ is the number of generated tokens in trajectory $y_i$, and $D_{\mathrm{KL}}$ is the token-level regularizer against a frozen reference policy. We recompute log probabilities before each update, use asymmetric clipping bounds, and filter groups whose rewards are identical, since they provide no relative learning signal. This objective directly couples the localization decision with the final answer reward, while the value-boost rollout engine ensures that rewards are assigned to trajectories that actually use LoRe during generation.

\begin{figure}[t]
    \centering
    \includegraphics[width=\linewidth]{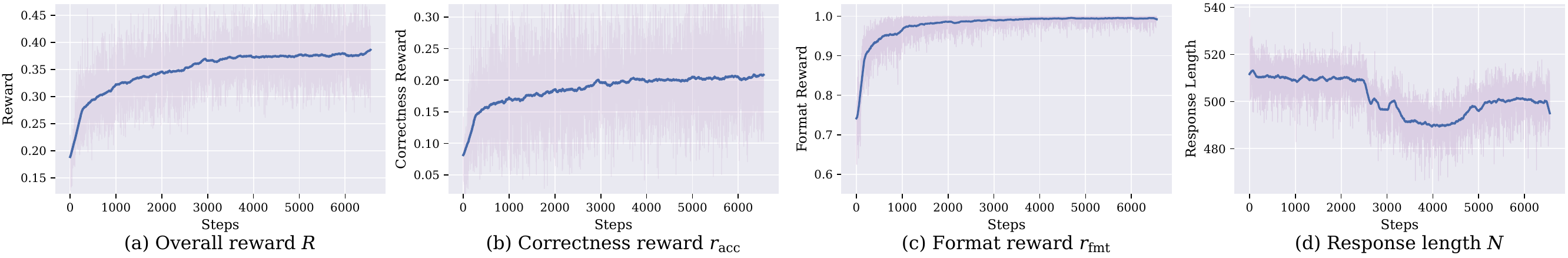}
    \vspace{-0.6cm}
    \caption{\textbf{GRPO training dynamics for LoRe with Qwen2.5-VL-7B~\cite{bai2025qwen25vl}.} The increasing correctness reward with stable response length indicates that RL improves the LoRe's reasoning ability.}
    \vspace{-0.2cm}
    \label{fig:qwen25_training_curves}
\end{figure}

Figure~\ref{fig:qwen25_training_curves} shows the training curves for Qwen2.5-VL-7B~\cite{bai2025qwen25vl}. The overall reward and correctness reward steadily increase, indicating that GRPO improves task-level reasoning rather than only fitting the required response format. The format reward saturates early, which suggests that the model quickly learns to produce parseable spans for the rollout engine. Meanwhile, the response length remains stable, indicating that LoRe better utilizes visual evidence and learns to reason efficiently.

\subsection{COCO-ObjQA benchmark}
\label{sec:data}

LoRe focuses on cases where the answer is visible but requires selecting a relevant object or attribute. We construct COCO-ObjQA from MS COCO~\cite{lin2014microsoft} instance annotations to evaluate this ability across target sizes, complementing the smaller V* benchmark~\cite{wu2024vstar} and resolution-focused HR-Bench~\cite{wang2025divide}. We retain targets whose category appears once in the image to reduce instance ambiguity. Gemini 3~\cite{pichai2025gemini3} receives the full image, target crop, category, and box to generate a multiple-choice question. Validation includes structural checks, independent GPT-5.4 checks of answerability and label correctness from the full image, and targeted manual inspection of high-risk samples. Confirmed failures are removed. Evaluation models receive only the original full image and question, including its answer options, without boxes, crops, or highlighted regions. COCO-ObjQA contains 8,537 QA pairs from 4,107 images across all 80 COCO categories. By target-area ratio (COCO box area divided by image area), 1,743 examples are small ($\leq$1\%), 3,036 medium (1--10\%), and 3,758 large ($>$10\%). This is a controlled diagnostic of object-centric evidence selection, not a comprehensive test of multi-object relations or image transformations.

\begin{figure}[t]
    \centering
    \includegraphics[width=1\linewidth]{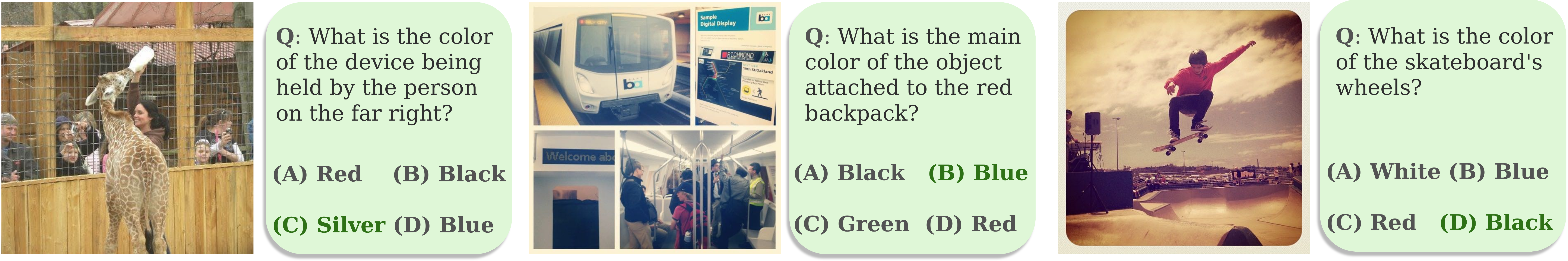}
    \vspace{-0.6cm}
    \caption{\textbf{Qualitative examples from COCO-ObjQA.} Boxes visualize target COCO annotations across object scales and are not part of the evaluation input. Models receive the original full image and the multiple-choice question.}
    \label{fig:coco_objqa_examples}
    \vspace{-0.5cm}
\end{figure}

\section{Experiments}
\label{sec:exp}

\subsection{Experimental setting}
\label{sec:exp_setting}

\textbf{Benchmarks and metrics.} We evaluate LoRe on three held-out benchmarks that probe complementary forms of fine-grained visual reasoning. V*~\cite{wu2024vstar} benchmark focuses on high-resolution visual search, including attribute recognition and relative-position reasoning, while HR-Bench~\cite{wang2025divide} stresses perception under 4K and 8K image resolutions. COCO-ObjQA, introduced in Section~\ref{sec:data}, provides the main object-level diagnostic: each question is tied to a known target instance and is answered from the original full image, allowing us to test whether a method can find and use localized evidence rather than rely on global image priors. We report exact-match accuracy by extracting the final option from each model response. For COCO-ObjQA, we also report accuracy by target-object size, with the small-object split most directly reflecting LoRe's evidence-selection ability.

\textbf{Training details.} We train LoRe using the GRPO objective and rollout procedure described in Section~\ref{sec:training}. The training data contains Ref-L4~\cite{chen2025revisiting}, ViGoRL-SA and SAT-2~\cite{sarch2025vigorl}, PixMo-Count~\cite{deitke2025molmo}, InfographicVQA~\cite{mathew2022infographicvqa}, and AI2D~\cite{kembhavi2016diagram}, which provide supervision signals for localization, spatial reasoning, counting, OCR, and diagram understanding. We initialize the policy from Qwen2.5-VL-7B~\cite{bai2025qwen25vl} or Qwen3-VL-8B~\cite{bai2025qwen3} and train for roughly 7K steps on 8 NVIDIA H200 GPUs, which takes about 65 hours. Rollout generation uses our customized SGLang~\cite{zheng2024sglang} implementation of LoRe, while policy optimization is performed with FSDP. In our default setup, four GPUs serve rollouts, and four GPUs compute actor and reference-policy log probabilities. Unless otherwise stated, we use bfloat16 precision with a batch size of 32, and a maximum response length of 4096. We use Adam with a learning rate of $1\times10^{-6}$, a KL coefficient of 0.01, and the value boosting factor $\gamma$ of 2.

\textbf{Comparison protocol.} DeepEyes, ViLaSR, and Thyme are evaluated zero-shot using their official checkpoints under our common evaluation protocol. Their training data and objectives differ from LoRe's. Table~\ref{tab:sota} therefore compares complete systems rather than isolating the cache intervention under matched training. Latency is measured on the same H200 machine with batch size one, without an external inference engine, and includes each method's visual operations and VLM interactions. For LoRe, this also includes region parsing and KV cache updates. Further details are in Appendix~\ref{app:evaluation_details}.

\subsection{Main results}
\label{sec:main_results}

\begin{figure}[t]
    \centering
    \includegraphics[width=\linewidth]{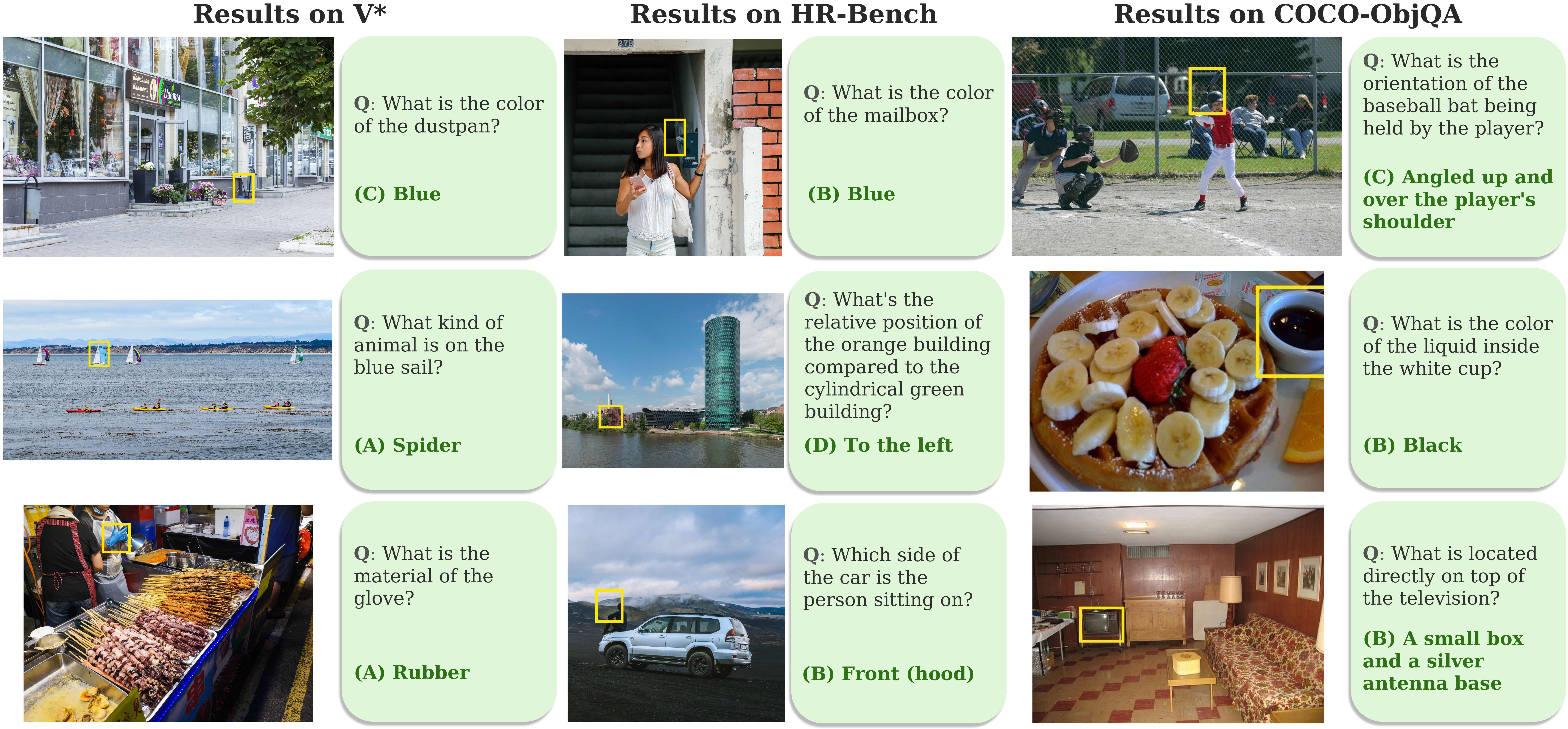}
    \vspace{-0.4cm}
    \caption{\textbf{Qualitative effects of LoRe.} LoRe localizes task-relevant visual evidence across different benchmarks before producing the final answer. Yellow boxes denote the region detection results.}
    \label{fig:qualitative_vis}
    \vspace{-0.4cm}
\end{figure}

Figure~\ref{fig:qualitative_vis} shows qualitative results of LoRe across different benchmarks, while Table~\ref{tab:sota} summarizes the quantitative comparison. We compare different thinking algorithms on V*~\cite{wu2024vstar}, HR-Bench~\cite{wang2025divide}, and COCO-ObjQA, and report latency to quantify the inference overhead.

\begin{table*}[t]
    \centering
    \caption{\textbf{Main comparison} on fine-grained multimodal reasoning benchmarks. Explicit reasoning baselines use official checkpoints with different training recipes. Gain rows report speedup over the fastest explicit reasoning baseline and relative accuracy gains over the corresponding base VLM. Bold and underlined scores mark the best and second-best results.}
    \vspace{0.3cm}
    \label{tab:sota}
    \small
    \setlength{\tabcolsep}{4.2pt}
    \renewcommand{\arraystretch}{1.16}
    \definecolor{loreRule}{RGB}{188,172,160}
    \definecolor{loreHL}{RGB}{255,242,235}
    \definecolor{loreDeltaHL}{RGB}{255,248,244}
    \definecolor{loreDelta}{RGB}{178,75,48}
    \begingroup
    \arrayrulecolor{loreRule}
    \resizebox{\textwidth}{!}{%
    \begin{tabular}{l c ccc ccc cccc}
        \toprule
        \textbf{Method} & \textbf{Latency} &
        \multicolumn{3}{c}{\textbf{V*}} &
        \multicolumn{3}{c}{\textbf{HR-Bench}} &
        \multicolumn{4}{c}{\textbf{COCO-ObjQA}} \\
        \cmidrule(lr){3-5}
        \cmidrule(lr){6-8}
        \cmidrule(lr){9-12}
        & & Dir. Attr. & Rel. Pos. & Avg. & 4K & 8K & Avg. & Small & Medium & Large & Avg. \\
        \midrule
        \multicolumn{12}{c}{\textit{Representative VLMs}} \\
        \midrule
        InternVL3-8B~\cite{zhu2025internvl3} & 1.25s & 72.1 & 68.3 & 70.6 & 62.7 & 52.4 & 57.6 & 79.8 & 91.7 & 92.2 & 89.5 \\
        Qwen2.5-VL-7B~\cite{bai2025qwen25vl} & 1.92s & 75.2 & 68.1 & 72.4 & 69.1 & 62.8 & 66.0 & 72.3 & 90.1 & 91.2 & 86.9 \\
        Qwen3-VL-8B~\cite{bai2025qwen3} & 2.95s & 82.4 & 80.9 & 81.8 & \underline{78.6} & \underline{74.1} & \underline{76.4} & 79.2 & 92.3 & 93.1 & 90.0 \\
        \midrule
        \multicolumn{12}{c}{\textit{State-of-the-art Reasoning Approaches}} \\
        \midrule
        ViLaSR-7B~\cite{wu2025reinforce} & 6.05s & 46.3 & 69.7 & 55.6 & 60.8 & 55.7 & 58.3 & 35.4 & 55.1 & 69.7 & 57.5 \\
        DeepEyes-7B~\cite{zheng2025deepeyes} & 34.73s & \textbf{91.1} & \underline{87.8} & \textbf{89.8} & 75.2 & 72.9 & 74.1 & 82.6 & 91.4 & 94.0 & 90.7 \\
        Thyme-7B~\cite{zhang2025thyme} & 6.18s & 82.7 & 80.1 & 81.7 & 77.9 & 72.2 & 75.1 & 80.2 & 91.2 & 93.0 & 89.7 \\
        \midrule
        \multicolumn{12}{c}{\textit{Ours}} \\
        \midrule
        \rowcolor{loreHL}
        LoRe (Qwen2.5-VL-7B) & 2.16s & 86.1 & 77.2 & 82.6 & 76.2 & 71.0 & 73.6 & \underline{83.1} & \underline{93.2} & \underline{94.2} & \underline{91.6} \\
        \rowcolor{loreDeltaHL}
        Gain / speedup & \textcolor{loreDelta}{2.8$\times$} & \textcolor{loreDelta}{+14.5\%} & \textcolor{loreDelta}{+13.4\%} & \textcolor{loreDelta}{+14.1\%} & \textcolor{loreDelta}{+10.3\%} & \textcolor{loreDelta}{+13.1\%} & \textcolor{loreDelta}{+11.5\%} & \textcolor{loreDelta}{+14.9\%} & \textcolor{loreDelta}{+3.4\%} & \textcolor{loreDelta}{+3.3\%} & \textcolor{loreDelta}{+5.4\%} \\
        \rowcolor{loreHL}
        LoRe (Qwen3-VL-8B) & 3.06s & \underline{88.8} & \textbf{88.2} & \underline{88.6} & \textbf{82.1} & \textbf{78.3} & \textbf{80.2} & \textbf{84.3} & \textbf{94.5} & \textbf{95.1} & \textbf{92.7} \\
        \rowcolor{loreDeltaHL}
        Gain / speedup & \textcolor{loreDelta}{2.0$\times$} & \textcolor{loreDelta}{+7.8\%} & \textcolor{loreDelta}{+9.0\%} & \textcolor{loreDelta}{+8.3\%} & \textcolor{loreDelta}{+4.5\%} & \textcolor{loreDelta}{+5.7\%} & \textcolor{loreDelta}{+5.0\%} & \textcolor{loreDelta}{+6.4\%} & \textcolor{loreDelta}{+2.4\%} & \textcolor{loreDelta}{+2.1\%} & \textcolor{loreDelta}{+3.0\%} \\
        \bottomrule
    \end{tabular}
    }
    \arrayrulecolor{black}
    \endgroup
    \vspace{-0.6cm}
\end{table*}

\noindent\textbf{Representative VLMs still have limited fine-grained reasoning ability.}
Table~\ref{tab:sota} first shows that direct VLM inference remains limited on fine-grained visual reasoning, even when the backbone is relatively strong. Among representative VLMs, Qwen3-VL-8B performs the best, but it reaches only 81.8\% average accuracy on V* and 76.4\% on HR-Bench. COCO-ObjQA further makes the object-scale failure mode explicit. Although Qwen3-VL-8B reaches 90.0\% average accuracy on COCO-ObjQA, it obtains only 79.2\% on small objects, compared with 92.3\% and 93.1\% on medium and large objects. The same pattern is stronger for Qwen2.5-VL-7B, which drops to 72.3\% on small objects while achieving 90.1\% and 91.2\% on medium and large objects.

\noindent\textbf{Explicit reasoning offers a different accuracy--latency tradeoff.}
DeepEyes-7B achieves the highest V* average, 89.8\%, at 34.73s per sample. ViLaSR-7B is faster at 6.05s but obtains 55.6\% on V* and 57.5\% on COCO-ObjQA under our protocol. These scores describe the transferred checkpoint, not the capability of crop-based reasoning under matched training. Thyme-7B achieves 75.1\% on HR-Bench and 89.7\% on COCO-ObjQA at 6.18s. These results show that explicit visual processing can improve accuracy, with benefits that depend on the checkpoint and benchmark, while requiring additional inference overhead.

\noindent\textbf{LoRe improves reasoning while preserving efficient inference.}
LoRe provides a different route to reasoning gains by turning visual focusing into a cache-side operation within a single interaction turn. With Qwen2.5-VL-7B as the backbone, LoRe improves the V* average from 72.4\% to 82.6\%, HR-Bench from 66.0\% to 73.6\%, and COCO-ObjQA from 86.9\% to 91.6\%. The largest COCO-ObjQA gain appears on small objects, where accuracy increases from 72.3\% to 83.1\%. With Qwen3-VL-8B, LoRe further reaches the best HR-Bench and COCO-ObjQA results, achieving 80.2\% and 92.7\% average accuracy, respectively, while also obtaining the second-best V* average at 88.6\%. Crucially, these gains do not require explicit image manipulation or an additional VLM call: LoRe (Qwen2.5-VL-7B) runs at 2.16s and is 2.8$\times$ faster than the fastest explicit reasoning baseline, while LoRe (Qwen3-VL-8B) remains 2.0$\times$ faster. These results show that LoRe improves fine-grained reasoning by reusing existing visual-token caches, avoiding the extra latency of tool-based visual reasoning.

\noindent\textbf{Remaining V* performance gap.} On Direct Attribute, LoRe (Qwen2.5-VL-7B) scores 86.1\% versus DeepEyes' 91.1\%, a 5.0-point gap. The larger 10.6-point gap, 77.2\% versus 87.8\%, is on Relative Position. Explicit cropping and re-encoding can expose fine detail absent from the initial representation, while a single predicted box may omit a reference object, select a distractor, or inadequately cover distant objects. These limitations help explain the complementary tradeoff: LoRe improves Relative Position over its base model's 68.1\%, while running approximately 16$\times$ faster than DeepEyes overall.

\subsection{Ablation study}
\label{sec:ablation}

We study the training-free pipeline, subsequent RL training, reward terms, amplification strength, and number of focused regions using Qwen2.5-VL-7B and the evaluation protocol of Section~\ref{sec:main_results}.


\begin{table*}[t]
    \centering
    \caption{\textbf{Training-free LoRe, RL training, and reward ablations.} LoRe variants use $\gamma=2$, while the base model uses no modulation. These rows do not constitute a fixed-checkpoint ablation of modulation after RL. COCO-ObjQA$_{\mathrm{IoU}}$ is mean box IoU, with invalid predictions counted as zero.}
    \vspace{0.3cm}
    \label{tab:ablation_training}
    \small
    \setlength{\tabcolsep}{5.5pt}
    \renewcommand{\arraystretch}{1.18}
    \definecolor{ablRule}{RGB}{190,178,166}
    \definecolor{ablHL}{RGB}{255,243,235}
    \begingroup
    \arrayrulecolor{ablRule}
    \resizebox{\textwidth}{!}{%
    \begin{tabular}{l ccccc}
        \toprule
        \textbf{Ablation model variant} & \textbf{V*$_{\mathrm{avg}}$} & \textbf{HR-Bench$_{\mathrm{avg}}$} & \textbf{COCO-ObjQA$_{\mathrm{small}}$} & \textbf{COCO-ObjQA$_{\mathrm{avg}}$} & \textbf{COCO-ObjQA$_{\mathrm{IoU}}$} \\
        \midrule
        \multicolumn{6}{c}{\textit{Implicit reasoning ablations}} \\
        \midrule
        Qwen2.5-VL-7B base & 72.4 & 66.0 & 72.3 & 86.9 & 0.24 \\
        + implicit thinking, w/o RL & 77.4 & 70.2 & 78.1 & 89.4 & 0.31 \\
        \rowcolor{ablHL}
        + implicit thinking and RL (LoRe) & \textbf{82.6} & \textbf{73.6} & \textbf{83.1} & \textbf{91.6} & \textbf{0.43} \\
        \midrule
        \multicolumn{6}{c}{\textit{Reward-term ablations}} \\
        \midrule
        \rowcolor{ablHL}
        Full reward & \textbf{82.6} & \textbf{73.6} & \textbf{83.1} & \textbf{91.6} & \textbf{0.43} \\
        w/o answer reward $r_{\mathrm{acc}}$ & 70.1 & 65.3 & 70.2 & 84.8 & 0.40 \\
        w/o answer-format reward $r_{\mathrm{fmt}}$ & 80.1 & 72.8 & 81.2 & 90.5 & 0.39 \\
        w/o grounding reward $r_{\mathrm{loc}}$ & 78.2 & 70.8 & 78.4 & 89.2 & 0.28 \\
        \bottomrule
    \end{tabular}
    }
    \arrayrulecolor{black}
    \endgroup
    \vspace{-0.3cm}
\end{table*}

\noindent\textbf{Training-free focusing and subsequent RL.}~Table~\ref{tab:ablation_training} compares the base VLM, training-free LoRe, and LoRe after RL. The training-free pipeline improves V* from 72.4\% to 77.4\%, HR-Bench from 66.0\% to 70.2\%, and COCO-ObjQA small objects from 72.3\% to 78.1\%. RL raises these scores to 82.6\%, 73.6\%, and 83.1\%, respectively, and the COCO-ObjQA average from 89.4\% to 91.6\%. Mean COCO-ObjQA IoU increases from 0.24 to 0.31 and then 0.43. These comparisons show gains from the training-free pipeline and from training within it, but they do not isolate test-time modulation from RL using a fixed trained checkpoint.

\noindent\textbf{Contribution of each reward term.}~Removing $r_{\mathrm{acc}}$ in Table~\ref{tab:ablation_training} causes the largest accuracy drop, reducing V* from 82.6\% to 70.1\% and COCO-ObjQA from 91.6\% to 84.8\%, supporting answer correctness as the main learning signal. Removing $r_{\mathrm{fmt}}$ causes smaller, consistent losses. Without $r_{\mathrm{loc}}$, mean COCO-ObjQA IoU falls from 0.43 to 0.28, while accuracy drops by 4.4, 2.8, and 2.4 percentage points on V*, HR-Bench, and COCO-ObjQA. Thus, grounding supervision benefits both localization and final answers, beyond learning the answer format alone.

\begin{figure}[t]
    \centering
    \definecolor{kvFocusRule}{RGB}{190,178,166}
    \definecolor{kvFocusHL}{RGB}{255,243,235}
    \begin{minipage}{0.50\linewidth}
        \centering
        \includegraphics[width=\linewidth]{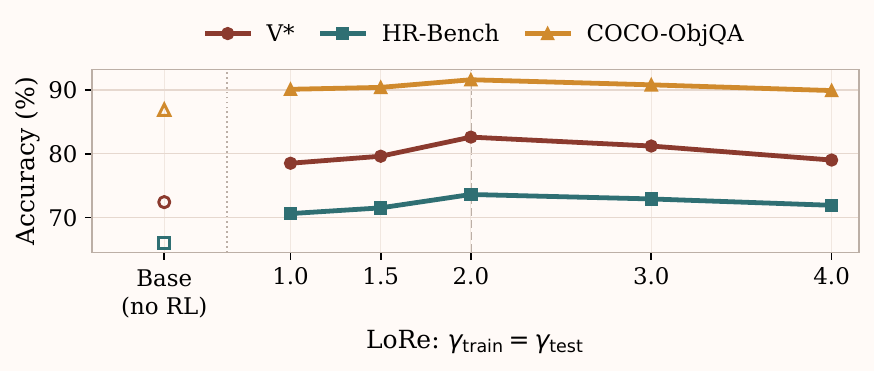}
    \end{minipage}
    \hfill
    \begin{minipage}{0.49\linewidth}
        \centering
        \footnotesize
        \setlength{\tabcolsep}{2.3pt}
        \renewcommand{\arraystretch}{1.08}
        \arrayrulecolor{kvFocusRule}
        \textit{Number of focused regions}\par\vspace{3pt}
        \begin{tabular}{l ccc}
            \toprule
            \textbf{Setting} & \textbf{V*$_{\rm avg}$} & \textbf{HR-Bench$_{\rm avg}$} & \textbf{COCO-ObjQA$_{\rm avg}$} \\
            \midrule
            Base (no RL) & 72.4 & 66.0 & 86.9 \\
            \rowcolor{kvFocusHL}
            $K=1$ & \textbf{82.6} & \textbf{73.6} & \textbf{91.6} \\
            $K=2$ & 81.5 & 73.6 & 91.3 \\
            $K=3$ & 80.2 & 72.9 & 90.5 \\
            Oracle box & 84.1 & 75.8 & 93.0 \\
            \bottomrule
        \end{tabular}
        \arrayrulecolor{black}
    \end{minipage}
    \caption{\textbf{KV cache modulation configurations.} Left: LoRe policies are trained and evaluated with the same indicated $\gamma$. The non-RL base reference is displayed separately. Right: one region performs best overall among learned settings. The oracle row substitutes reference boxes using the same RL checkpoint and $\gamma=2$. This comparison highlights the potential for more accurate region localization to further improve reasoning performance of LoRe.}
    \label{fig:ablation_kv_focus}
    \vspace{-0.1cm}
\end{figure}

\noindent\textbf{KV cache modulation strategies.} In the left panel of Figure~\ref{fig:ablation_kv_focus}, each LoRe point uses a policy trained and evaluated with the indicated factor, $\gamma_{\mathrm{train}}=\gamma_{\mathrm{test}}$. The $\gamma=1$ point uses RL without cache modulation during training or inference. The non-RL base reference is shown separately. This comparison therefore varies the training and inference configurations jointly. Among the evaluated LoRe settings, $\gamma=2$ gives the best accuracy: 82.6\% on V*, 73.6\% on HR-Bench, and 91.6\% on COCO-ObjQA. Larger factors reduce accuracy, consistent with excessive local amplification weakening the relative contribution of global context.

The right panel compares one, two, and three independently predicted regions, alongside the non-RL base reference. A single region gives the best learned result overall: V* accuracy decreases from 82.6\% to 81.5\% and 80.2\% with two and three regions. Two regions match one region on HR-Bench but perform worse on COCO-ObjQA. Additional regions may introduce distractors rather than useful evidence. A single enclosing box can cover multiple objects, but cannot weight them separately and may include irrelevant context between distant objects.

For the oracle-box comparison, we keep the RL checkpoint, $\gamma=2$, and decoding procedure fixed and replace predicted regions with COCO annotations on COCO-ObjQA or SAM2-derived~\cite{ravi2025sam2} pseudo-ground-truth regions on V* and HR-Bench. The respective gains are 1.4, 1.5, and 2.2 percentage points. This comparison with reference regions suggests that more accurate localization could further improve LoRe's reasoning performance. It is not a strict upper bound, especially when reference boxes are approximate, and does not attribute all remaining errors to localization.

\begin{table}[!htbp]
    \centering
    \caption{\textbf{Controlled comparisons with Qwen2.5-VL-7B.} The two $\gamma_{\mathrm{train}}=2$ rows share a checkpoint. Crop-based GRPO matches LoRe's backbone, training data, recipe, and objectives. All scores are average accuracy (\%).}
    \label{tab:ablation_controls}
    \small
    \setlength{\tabcolsep}{6pt}
    \renewcommand{\arraystretch}{1.12}
    \begin{tabular}{@{}lccccc@{}}
        \toprule
        \textbf{Policy} & $\gamma_{\mathrm{train}}$ & $\gamma_{\mathrm{test}}$ & \textbf{V*} & \textbf{HR-Bench} & \textbf{COCO-ObjQA} \\
        \midrule
        RL, no modulation & 1 & 1 & 78.5 & 70.6 & 90.1 \\
        LoRe, modulation disabled & 2 & 1 & 76.9 & 69.8 & 90.6 \\
        LoRe, default & 2 & 2 & \textbf{82.6} & \textbf{73.6} & 91.6 \\
        Crop-based GRPO & -- & -- & 80.9 & 71.8 & \textbf{92.3} \\
        \bottomrule
    \end{tabular}
\end{table}

\noindent\textbf{Controlling for RL and training data.} Table~\ref{tab:ablation_controls} separates the effects of training and test-time modulation. Training and testing at $\gamma=1$ improve over the base VLM, confirming gains from RL alone. For the checkpoint trained at $\gamma=2$, disabling modulation preserves the box output, prompt, and decoding procedure. Restoring modulation improves V*, HR-Bench, and COCO-ObjQA accuracy by 5.7, 3.8, and 1.0 percentage points, respectively. We also train a crop-based GRPO policy with the same backbone, data, recipe, and objectives. It predicts one box and re-encodes the crop as a second visual observation before answering. LoRe exceeds this control by 1.7 points on V* and 1.8 on HR-Bench, while trailing by 0.7 on COCO-ObjQA, without additional visual encoding. Appendix~\ref{app:additional_experiments} provides further modulation, fallback, localization, and transfer analyses.

\noindent\textbf{Localization reliability.} RL reduces fallback rates from 11.0\% to 6.8\% on V*, 8.5\% to 5.0\% on HR-Bench, and 2.5\% to 1.5\% on COCO-ObjQA (Appendix~\ref{app:fallback_results}). At $\gamma=2$, oracle boxes correct 20.7\%, 16.3\%, and 38.1\% of predicted-box errors, respectively (Appendix~\ref{app:oracle_analysis}). Remaining errors indicate that accurate localization alone does not guarantee correct reasoning.

\section{Conclusion}
\label{sec:conclusion}
We present LoRe, a framework for single-turn visual focusing and multimodal reasoning. The VLM emits a question-relevant region, and LoRe amplifies its visual value-cache entries before generation continues in the same request. GRPO training couples region selection with answer correctness, response format, and visual grounding. COCO-ObjQA complements V* and HR-Bench by enabling object-centric evaluation across target sizes. Across these benchmarks, LoRe improves the corresponding base VLMs while adding little inference overhead, offering a useful accuracy--latency tradeoff relative to explicit visual-tool methods. Controlled comparisons further show that cache modulation contributes beyond the gains from RL alone.

\noindent\textbf{Limitations and future work.} LoRe amplifies encoded evidence, but it cannot recover discarded detail or replace operations that transform and re-encode an image. A single region may omit relevant objects or include distractors between distant objects, and multiple independent regions do not consistently improve the reported results. COCO-ObjQA emphasizes single-target evidence selection rather than comprehensive relational reasoning. More reliable multi-object localization and selective use of crop or zoom tools are promising extensions, but a hybrid routing mechanism is outside the present single-turn setting.

\begin{ack}
We gratefully acknowledge Prof. Stefanos Zafeiriou for his valuable insights and discussions on tiny object perception with multimodal large language models. This work was supported by the Turing AI Fellowship
(EP\slash Z534699\slash 1). Fei Ni was supported by a MSCA Postdoctoral Fellowship. Jiankang Deng was supported by the NVIDIA Academic Grant. The authors gratefully acknowledge the computing resources provided by Isambard-AI, part of the National AI Research Resource (AIRR).
\end{ack}

\clearpage
{
    \small
    \bibliographystyle{ieeenat_fullname}
    \bibliography{main}

\begin{thebibliography}{59}
\providecommand{\natexlab}[1]{#1}
\providecommand{\url}[1]{\texttt{#1}}
\expandafter\ifx\csname urlstyle\endcsname\relax
  \providecommand{\doi}[1]{doi: #1}\else
  \providecommand{\doi}{doi: \begingroup \urlstyle{rm}\Url}\fi

\bibitem[Alayrac et~al.(2022)Alayrac, Donahue, Luc, Miech, Barr, Hasson, Lenc,
  Mensch, Millican, Reynolds, Ring, Rutherford, Cabi, Han, Gong, Samangooei,
  Monteiro, Menick, Borgeaud, Brock, Nematzadeh, Sharifzadeh, Binkowski,
  Barreira, Vinyals, Zisserman, and Simonyan]{Alayracetal2022}
Jean-Baptiste Alayrac, Jeff Donahue, Pauline Luc, Antoine Miech, Iain Barr,
  Yana Hasson, Karel Lenc, Arthur Mensch, Katie Millican, Malcolm Reynolds,
  Roman Ring, Eliza Rutherford, Serkan Cabi, Tengda Han, Zhitao Gong, Sina
  Samangooei, Marianne Monteiro, Jacob Menick, Sebastian Borgeaud, Andrew
  Brock, Aida Nematzadeh, Sahand Sharifzadeh, Mikolaj Binkowski, Ricardo
  Barreira, Oriol Vinyals, Andrew Zisserman, and Karen Simonyan.
\newblock Flamingo: a visual language model for few-shot learning.
\newblock In \emph{NeurIPS}, 2022.

\bibitem[Bai et~al.(2023)Bai, Bai, Chu, Cui, Dang, Deng, Fan, Ge, Han, Huang,
  Hui, et~al.]{bai2023qwentechnicalreport}
Jinze Bai, Shuai Bai, Yunfei Chu, Zeyu Cui, Kai Dang, Xiaodong Deng, Yang Fan,
  Wenbin Ge, Yu Han, Fei Huang, Binyuan Hui, et~al.
\newblock Qwen technical report.
\newblock \emph{arXiv:2309.16609}, 2023.

\bibitem[Bai et~al.(2025{\natexlab{a}})Bai, Cai, Chen, Chen, Chen, Cheng, Deng,
  Ding, Gao, Ge, et~al.]{bai2025qwen3}
Shuai Bai, Yuxuan Cai, Ruizhe Chen, Keqin Chen, Xionghui Chen, Zesen Cheng,
  Lianghao Deng, Wei Ding, Chang Gao, Chunjiang Ge, et~al.
\newblock {Qwen3-VL} technical report.
\newblock \emph{arXiv:2511.21631}, 2025{\natexlab{a}}.

\bibitem[Bai et~al.(2025{\natexlab{b}})Bai, Chen, Liu, Wang, Ge, Song, Dang,
  Wang, Wang, Tang, Zhong, Zhu, Yang, Li, Wan, Wang, Ding, Fu, Xu, Ye, Zhang,
  Xie, Cheng, Zhang, Yang, Xu, and Lin]{bai2025qwen25vl}
Shuai Bai, Keqin Chen, Xuejing Liu, Jialin Wang, Wenbin Ge, Sibo Song, Kai
  Dang, Peng Wang, Shijie Wang, Jun Tang, Humen Zhong, Yuanzhi Zhu, Mingkun
  Yang, Zhaohai Li, Jianqiang Wan, Pengfei Wang, Wei Ding, Zheren Fu, Yiheng
  Xu, Jiabo Ye, Xi Zhang, Tianbao Xie, Zesen Cheng, Hang Zhang, Zhibo Yang,
  Haiyang Xu, and Junyang Lin.
\newblock Qwen2.5-vl technical report.
\newblock \emph{arXiv:2502.13923}, 2025{\natexlab{b}}.

\bibitem[Chen et~al.(2025{\natexlab{a}})Chen, Wei, Zhao, Song, Wu, Peng, Chan,
  and Zhang]{chen2025revisiting}
Jierun Chen, Fangyun Wei, Jinjing Zhao, Sizhe Song, Bohuai Wu, Zhuoxuan Peng,
  S.-H.~Gary Chan, and Hongyang Zhang.
\newblock Revisiting referring expression comprehension evaluation in the era
  of large multimodal models.
\newblock In \emph{CVPR Workshops}, 2025{\natexlab{a}}.

\bibitem[Chen et~al.(2025{\natexlab{b}})Chen, Zhang, Huang, Niu, Zhang, Wen,
  and Hu]{chen2025ict}
Junzhe Chen, Tianshu Zhang, Shiyu Huang, Yuwei Niu, Linfeng Zhang, Lijie Wen,
  and Xuming Hu.
\newblock {ICT}: Image-object cross-level trusted intervention for mitigating
  object hallucination in large vision-language models.
\newblock In \emph{Proceedings of the IEEE/CVF Conference on Computer Vision
  and Pattern Recognition}, pages 4209--4221, 2025{\natexlab{b}}.

\bibitem[Chen et~al.(2024)Chen, Sikka, Cogswell, Ji, and
  Divakaran]{chen2024measuring}
Yangyi Chen, Karan Sikka, Michael Cogswell, Heng Ji, and Ajay Divakaran.
\newblock Measuring and improving chain-of-thought reasoning in vision-language
  models.
\newblock In \emph{NAACL}, 2024.

\bibitem[Chern et~al.(2025)Chern, Hu, Chern, Kou, Su, Ma, Deng, and
  Liu]{chern2025thinkinggeneratedimages}
Ethan Chern, Zhulin Hu, Steffi Chern, Siqi Kou, Jiadi Su, Yan Ma, Zhijie Deng,
  and Pengfei Liu.
\newblock Thinking with generated images.
\newblock \emph{arXiv:2505.22525}, 2025.

\bibitem[Dahou et~al.(2025)Dahou, Huynh, Le-Khac, Para, Singh, and
  Narayan]{huynh2025vision}
Yasser Dahou, Ngoc~Dung Huynh, Phuc~H Le-Khac, Wamiq~Reyaz Para, Ankit Singh,
  and Sanath Narayan.
\newblock Vision-language models can't see the obvious.
\newblock In \emph{ICCV}, 2025.

\bibitem[{DeepSeek-AI} et~al.(2024){DeepSeek-AI}, Liu, Feng, Xue, Wang, Wu, Lu,
  Zhao, Deng, Zhang, Ruan, et~al.]{liu2024deepseek}
{DeepSeek-AI}, Aixin Liu, Bei Feng, Bing Xue, Bingxuan Wang, Bochao Wu, Chengda
  Lu, Chenggang Zhao, Chengqi Deng, Chenyu Zhang, Chong Ruan, et~al.
\newblock {DeepSeek-V3} technical report.
\newblock \emph{arXiv:2412.19437}, 2024.

\bibitem[{DeepSeek-AI} et~al.(2025){DeepSeek-AI}, Guo, Yang, Zhang, Song, Wang,
  Zhu, Xu, Zhang, Ma, Bi, et~al.]{guo2025deepseek}
{DeepSeek-AI}, Daya Guo, Dejian Yang, Haowei Zhang, Junxiao Song, Peiyi Wang,
  Qihao Zhu, Runxin Xu, Ruoyu Zhang, Shirong Ma, Xiao Bi, et~al.
\newblock {DeepSeek-R1}: Incentivizing reasoning capability in {LLMs} via
  reinforcement learning.
\newblock \emph{arXiv:2501.12948}, 2025.

\bibitem[Deitke et~al.(2025)Deitke, Clark, Lee, Tripathi, Yang, Park, Salehi,
  Muennighoff, Lo, Soldaini, et~al.]{deitke2025molmo}
Matt Deitke, Christopher Clark, Sangho Lee, Rohun Tripathi, Yue Yang, Jae~Sung
  Park, Mohammadreza Salehi, Niklas Muennighoff, Kyle Lo, Luca Soldaini, et~al.
\newblock {Molmo} and {PixMo}: Open weights and open data for state-of-the-art
  vision-language models.
\newblock In \emph{CVPR}, 2025.

\bibitem[Dosovitskiy et~al.(2021)Dosovitskiy, Beyer, Kolesnikov, Weissenborn,
  Zhai, Unterthiner, Dehghani, Minderer, Heigold, Gelly,
  et~al.]{dosovitskiy2021image}
Alexey Dosovitskiy, Lucas Beyer, Alexander Kolesnikov, Dirk Weissenborn,
  Xiaohua Zhai, Thomas Unterthiner, Mostafa Dehghani, Matthias Minderer, Georg
  Heigold, Sylvain Gelly, et~al.
\newblock An image is worth 16x16 words: Transformers for image recognition at
  scale.
\newblock In \emph{ICLR}, 2021.

\bibitem[Fini et~al.(2025)Fini, Shukor, Li, Dufter, Klein, Haldimann,
  Aitharaju, da~Costa, B{\'e}thune, Gan, et~al.]{fini2025multimodal}
Enrico Fini, Mustafa Shukor, Xiujun Li, Philipp Dufter, Michal Klein, David
  Haldimann, Sai Aitharaju, Victor G~Turrisi da Costa, Louis B{\'e}thune, Zhe
  Gan, et~al.
\newblock Multimodal autoregressive pre-training of large vision encoders.
\newblock In \emph{CVPR}, 2025.

\bibitem[Gupta and Kembhavi(2023)]{gupta2023visual}
Tanmay Gupta and Aniruddha Kembhavi.
\newblock Visual programming: Compositional visual reasoning without training.
\newblock In \emph{CVPR}, 2023.

\bibitem[Kembhavi et~al.(2016)Kembhavi, Salvato, Kolve, Seo, Hajishirzi, and
  Farhadi]{kembhavi2016diagram}
Aniruddha Kembhavi, Mike Salvato, Eric Kolve, Minjoon Seo, Hannaneh Hajishirzi,
  and Ali Farhadi.
\newblock A diagram is worth a dozen images.
\newblock In \emph{ECCV}, 2016.

\bibitem[Kwon et~al.(2023)Kwon, Li, Zhuang, Sheng, Zheng, Yu, Gonzalez, Zhang,
  and Stoica]{kwon2023efficient}
Woosuk Kwon, Zhuohan Li, Siyuan Zhuang, Ying Sheng, Lianmin Zheng, Cody~Hao Yu,
  Joseph~E. Gonzalez, Hao Zhang, and Ion Stoica.
\newblock Efficient memory management for large language model serving with
  pagedattention.
\newblock In \emph{SOSP}, 2023.

\bibitem[Li et~al.(2024)Li, Zhang, Guo, Zhang, Li, Zhang, Zhang, Zhang, Li,
  Liu, et~al.]{li2024llava}
Bo Li, Yuanhan Zhang, Dong Guo, Renrui Zhang, Feng Li, Hao Zhang, Kaichen
  Zhang, Peiyuan Zhang, Yanwei Li, Ziwei Liu, et~al.
\newblock {LLaVA-OneVision}: Easy visual task transfer.
\newblock \emph{arXiv:2408.03326}, 2024.

\bibitem[Li et~al.(2025{\natexlab{a}})Li, Wu, Zhang, Xia, Mao, Dong, Vuli{\'c},
  and Wei]{li2025imagine}
Chengzu Li, Wenshan Wu, Huanyu Zhang, Yan Xia, Shaoguang Mao, Li Dong, Ivan
  Vuli{\'c}, and Furu Wei.
\newblock Imagine while reasoning in space: Multimodal
  visualization-of-thought.
\newblock In \emph{ICML}, 2025{\natexlab{a}}.

\bibitem[Li et~al.(2025{\natexlab{b}})Li, Xu, Zhao, and Peng]{li2025dyfo}
Geng Li, Jinglin Xu, Yunzhen Zhao, and Yuxin Peng.
\newblock {DyFo}: A training-free dynamic focus visual search for enhancing
  {LMMs} in fine-grained visual understanding.
\newblock In \emph{CVPR}, 2025{\natexlab{b}}.

\bibitem[Li et~al.(2022)Li, Li, Xiong, and Hoi]{Lietal2022}
Junnan Li, Dongxu Li, Caiming Xiong, and Steven Hoi.
\newblock {BLIP}: Bootstrapping language-image pre-training for unified
  vision-language understanding and generation.
\newblock In \emph{ICML}, 2022.

\bibitem[Li et~al.(2023)Li, Li, Savarese, and Hoi]{Lietal2023}
Junnan Li, Dongxu Li, Silvio Savarese, and Steven Hoi.
\newblock {BLIP}-2: Bootstrapping language-image pre-training with frozen image
  encoders and large language models.
\newblock In \emph{ICML}, 2023.

\bibitem[Li et~al.(2025{\natexlab{c}})Li, Zhong, Zhao, Zhang, Lin, Lai, Wei,
  Psounis, and Zhang]{li2025tirbench}
Ming Li, Jike Zhong, Shitian Zhao, Haoquan Zhang, Shaoheng Lin, Yuxiang Lai,
  Chen Wei, Konstantinos Psounis, and Kaipeng Zhang.
\newblock {TIR-Bench}: A comprehensive benchmark for agentic
  thinking-with-images reasoning.
\newblock \emph{arXiv:2511.01833}, 2025{\natexlab{c}}.

\bibitem[Lin et~al.(2014)Lin, Maire, Belongie, Hays, Perona, Ramanan,
  Doll{\'a}r, and Zitnick]{lin2014microsoft}
Tsung-Yi Lin, Michael Maire, Serge Belongie, James Hays, Pietro Perona, Deva
  Ramanan, Piotr Doll{\'a}r, and C~Lawrence Zitnick.
\newblock Microsoft {COCO}: Common objects in context.
\newblock In \emph{ECCV}, 2014.

\bibitem[Liu et~al.(2023)Liu, Li, Wu, and Lee]{liu2023visual}
Haotian Liu, Chunyuan Li, Qingyang Wu, and Yong~Jae Lee.
\newblock Visual instruction tuning.
\newblock In \emph{NeurIPS}, 2023.

\bibitem[Lu et~al.(2024)Lu, Bansal, Xia, Liu, Li, Hajishirzi, Cheng, Chang,
  Galley, and Gao]{lu2024mathvista}
Pan Lu, Hritik Bansal, Tony Xia, Jiacheng Liu, Chunyuan Li, Hannaneh
  Hajishirzi, Hao Cheng, Kai-Wei Chang, Michel Galley, and Jianfeng Gao.
\newblock {MathVista}: Evaluating mathematical reasoning of foundation models
  in visual contexts.
\newblock In \emph{International Conference on Learning Representations}, 2024.

\bibitem[Luo et~al.(2025)Luo, Zhou, Zhang, Zheng, Sun, and Ji]{luo2024feast}
Gen Luo, Yiyi Zhou, Yuxin Zhang, Xiawu Zheng, Xiaoshuai Sun, and Rongrong Ji.
\newblock Feast your eyes: Mixture-of-resolution adaptation for multimodal
  large language models.
\newblock In \emph{ICLR}, 2025.

\bibitem[Mathew et~al.(2022)Mathew, Bagal, Tito, Karatzas, Valveny, and
  Jawahar]{mathew2022infographicvqa}
Minesh Mathew, Viraj Bagal, Rub{\`e}n Tito, Dimosthenis Karatzas, Ernest
  Valveny, and C.~V. Jawahar.
\newblock {InfographicVQA}.
\newblock In \emph{WACV}, 2022.

\bibitem[{OpenAI}(2025)]{openai_o3_o4mini_system_card_2025}
{OpenAI}.
\newblock {OpenAI o3 and o4-mini System Card}, 2025.

\bibitem[Pichai et~al.(2025)Pichai, Hassabis, and
  Kavukcuoglu]{pichai2025gemini3}
Sundar Pichai, Demis Hassabis, and Koray Kavukcuoglu.
\newblock A new era of intelligence with {Gemini} 3.
\newblock
  \url{https://blog.google/products-and-platforms/products/gemini/gemini-3/},
  2025.
\newblock Accessed: 2026-05-07.

\bibitem[Qiao et~al.(2025)Qiao, Tan, Dong, Wu, Sun, Song, Wang, GongQue, Lei,
  Zhang, Wei, Zhang, Qiao, Zong, Xu, Yang, Bao, Diao, Li, and
  Zhang]{qiao2025wemath}
Runqi Qiao, Qiuna Tan, Guanting Dong, Minhui Wu, Chong Sun, Xiaoshuai Song,
  Jiapeng Wang, Zhuoma GongQue, Shanglin Lei, Yifan Zhang, Zhe Wei, Miaoxuan
  Zhang, Runfeng Qiao, Xiao Zong, Yida Xu, Peiqing Yang, Zhimin Bao, Muxi Diao,
  Chen Li, and Honggang Zhang.
\newblock {We-Math}: Does your large multimodal model achieve human-like
  mathematical reasoning?
\newblock In \emph{Proceedings of the 63rd Annual Meeting of the Association
  for Computational Linguistics (Volume 1: Long Papers)}, 2025.

\bibitem[Ravi et~al.(2025)Ravi, Gabeur, Hu, Hu, Ryali, Ma, Khedr, R{\"a}dle,
  Rolland, Gustafson, Mintun, Pan, Alwala, Carion, Wu, Girshick, Doll{\'a}r,
  and Feichtenhofer]{ravi2025sam2}
Nikhila Ravi, Valentin Gabeur, Yuan-Ting Hu, Ronghang Hu, Chaitanya Ryali,
  Tengyu Ma, Haitham Khedr, Roman R{\"a}dle, Chloe Rolland, Laura Gustafson,
  Eric Mintun, Junting Pan, Kalyan~Vasudev Alwala, Nicolas Carion, Chao-Yuan
  Wu, Ross Girshick, Piotr Doll{\'a}r, and Christoph Feichtenhofer.
\newblock {SAM 2}: Segment anything in images and videos.
\newblock In \emph{International Conference on Learning Representations}, 2025.

\bibitem[Sarch et~al.(2025)Sarch, Saha, Khandelwal, Jain, Tarr, Kumar, and
  Fragkiadaki]{sarch2025vigorl}
Gabriel Sarch, Snigdha Saha, Naitik Khandelwal, Ayush Jain, Michael~J. Tarr,
  Aviral Kumar, and Katerina Fragkiadaki.
\newblock Grounded reinforcement learning for visual reasoning.
\newblock In \emph{NeurIPS}, 2025.

\bibitem[Su et~al.(2025{\natexlab{a}})Su, Wang, Ren, Lin, and
  Chen]{wang2025pixel}
Alex Su, Haozhe Wang, Weiming Ren, Fangzhen Lin, and Wenhu Chen.
\newblock Pixel reasoner: Incentivizing pixel-space reasoning with
  curiosity-driven reinforcement learning.
\newblock In \emph{NeurIPS}, 2025{\natexlab{a}}.

\bibitem[Su et~al.(2025{\natexlab{b}})Su, Li, Song, Hao, Yang, Zhang, Chen, Gu,
  Li, Qu, et~al.]{su2025openthinkimg}
Zhaochen Su, Linjie Li, Mingyang Song, Yunzhuo Hao, Zhengyuan Yang, Jun Zhang,
  Guanjie Chen, Jiawei Gu, Juntao Li, Xiaoye Qu, et~al.
\newblock {OpenThinkIMG}: Learning to think with images via visual tool
  reinforcement learning.
\newblock \emph{arXiv:2505.08617}, 2025{\natexlab{b}}.

\bibitem[Su et~al.(2025{\natexlab{c}})Su, Xia, Guo, Liu, Ma, Qu, Liu, Li, Zeng,
  Yang, et~al.]{su2025thinking}
Zhaochen Su, Peng Xia, Hangyu Guo, Zhenhua Liu, Yan Ma, Xiaoye Qu, Jiaqi Liu,
  Yanshu Li, Kaide Zeng, Zhengyuan Yang, et~al.
\newblock Thinking with images for multimodal reasoning: Foundations, methods,
  and future frontiers.
\newblock \emph{arXiv:2506.23918}, 2025{\natexlab{c}}.

\bibitem[Sur{\'i}s et~al.(2023)Sur{\'i}s, Menon, and
  Vondrick]{suris2023vipergpt}
D{\'i}dac Sur{\'i}s, Sachit Menon, and Carl Vondrick.
\newblock {ViperGPT}: Visual inference via python execution for reasoning.
\newblock In \emph{ICCV}, 2023.

\bibitem[Sutton and Barto(1998)]{sutton1998reinforcement}
Richard~S Sutton and Andrew~G Barto.
\newblock \emph{Reinforcement learning: An introduction}.
\newblock MIT press Cambridge, 1998.

\bibitem[Touvron et~al.(2023)Touvron, Lavril, Izacard, Martinet, Lachaux,
  Lacroix, Rozière, Goyal, Hambro, Azhar, Rodriguez, Joulin, Grave, and
  Lample]{Touvronetal2023}
Hugo Touvron, Thibaut Lavril, Gautier Izacard, Xavier Martinet, Marie-Anne
  Lachaux, Timothée Lacroix, Baptiste Rozière, Naman Goyal, Eric Hambro,
  Faisal Azhar, Aurelien Rodriguez, Armand Joulin, Edouard Grave, and Guillaume
  Lample.
\newblock {LLaMA}: Open and efficient foundation language models.
\newblock \emph{arXiv:2302.13971}, 2023.

\bibitem[Tschannen et~al.(2025)Tschannen, Gritsenko, Wang, Naeem,
  Alabdulmohsin, Parthasarathy, Evans, Beyer, Xia, Mustafa,
  et~al.]{tschannen2025siglip}
Michael Tschannen, Alexey Gritsenko, Xiao Wang, Muhammad~Ferjad Naeem, Ibrahim
  Alabdulmohsin, Nikhil Parthasarathy, Talfan Evans, Lucas Beyer, Ye Xia, Basil
  Mustafa, et~al.
\newblock {SigLIP} 2: Multilingual vision-language encoders with improved
  semantic understanding, localization, and dense features.
\newblock \emph{arXiv:2502.14786}, 2025.

\bibitem[Vaswani et~al.(2017)Vaswani, Shazeer, Parmar, Uszkoreit, Jones, Gomez,
  Kaiser, and Polosukhin]{vaswani2017attention}
Ashish Vaswani, Noam Shazeer, Niki Parmar, Jakob Uszkoreit, Llion Jones,
  Aidan~N. Gomez, Lukasz Kaiser, and Illia Polosukhin.
\newblock Attention is all you need.
\newblock In \emph{NeurIPS}, 2017.

\bibitem[Wang et~al.(2025{\natexlab{a}})Wang, Ding, Zeng, Zhou, Shen, Luo, Yu,
  and Tao]{wang2025divide}
Wenbin Wang, Liang Ding, Minyan Zeng, Xiabin Zhou, Li Shen, Yong Luo, Wei Yu,
  and Dacheng Tao.
\newblock Divide, conquer and combine: A training-free framework for
  high-resolution image perception in multimodal large language models.
\newblock In \emph{AAAI}, 2025{\natexlab{a}}.

\bibitem[Wang et~al.(2025{\natexlab{b}})Wang, Gao, Gu, Pu, Cui, Wei, Liu, Jing,
  Ye, Shao, et~al.]{wang2025internvl3}
Weiyun Wang, Zhangwei Gao, Lixin Gu, Hengjun Pu, Long Cui, Xingguang Wei,
  Zhaoyang Liu, Linglin Jing, Shenglong Ye, Jie Shao, et~al.
\newblock {InternVL3.5}: Advancing open-source multimodal models in
  versatility, reasoning, and efficiency.
\newblock \emph{arXiv:2508.18265}, 2025{\natexlab{b}}.

\bibitem[Wang et~al.(2024)Wang, Zhou, Liu, Lu, Xu, He, Yoon, Lu, Liu,
  Bertasius, Bansal, Yao, and Huang]{wang2024mementos}
Xiyao Wang, Yuhang Zhou, Xiaoyu Liu, Hongjin Lu, Yuancheng Xu, Feihong He,
  Jaehong Yoon, Taixi Lu, Fuxiao Liu, Gedas Bertasius, Mohit Bansal, Huaxiu
  Yao, and Furong Huang.
\newblock Mementos: A comprehensive benchmark for multimodal large language
  model reasoning over image sequences.
\newblock In \emph{ACL}, 2024.

\bibitem[Wang et~al.(2025{\natexlab{c}})Wang, Wang, Cheng, Fei, Ding, Guo, Tao,
  and Qiu]{wang2025visuothink}
Yikun Wang, Siyin Wang, Qinyuan Cheng, Zhaoye Fei, Liang Ding, Qipeng Guo,
  Dacheng Tao, and Xipeng Qiu.
\newblock {VisuoThink}: Empowering {LVLM} reasoning with multimodal tree
  search.
\newblock In \emph{ACL}, 2025{\natexlab{c}}.

\bibitem[Wei et~al.(2022)Wei, Wang, Schuurmans, Bosma, Ichter, Xia, Chi, Le,
  and Zhou]{wei2022chain}
Jason Wei, Xuezhi Wang, Dale Schuurmans, Maarten Bosma, Brian Ichter, Fei Xia,
  Ed~H. Chi, Quoc~V Le, and Denny Zhou.
\newblock Chain-of-thought prompting elicits reasoning in large language
  models.
\newblock In \emph{NeurIPS}, 2022.

\bibitem[Wu et~al.(2025)Wu, Guan, Feng, Liu, Wu, Wang, Wu, and
  Tan]{wu2025reinforce}
Junfei Wu, Jian Guan, Kaituo Feng, Qiang Liu, Shu Wu, Liang Wang, Wei Wu, and
  Tieniu Tan.
\newblock Reinforcing spatial reasoning in vision-language models with
  interwoven thinking and visual drawing.
\newblock In \emph{NeurIPS}, 2025.

\bibitem[Wu et~al.(2026)Wu, Yang, Jiang, Li, Yan, Yu, Zhang, Zhai, and
  Nahrstedt]{wu2025vtool}
Mingyuan Wu, Jingcheng Yang, Jize Jiang, Meitang Li, Kaizhuo Yan, Hanchao Yu,
  Minjia Zhang, Chengxiang Zhai, and Klara Nahrstedt.
\newblock {VTool-R1}: {VLMs} learn to think with images via reinforcement
  learning on multimodal tool use.
\newblock In \emph{ICLR}, 2026.

\bibitem[Wu and Xie(2024)]{wu2024vstar}
Penghao Wu and Saining Xie.
\newblock V*: Guided visual search as a core mechanism in multimodal llms.
\newblock In \emph{CVPR}, 2024.

\bibitem[Xie et~al.(2025)Xie, Yang, An, Wu, Zhao, Deng, Ran, Wang, Feng, Miles,
  Elezi, and Deng]{yinxie_2025_rice}
Yin Xie, Kaicheng Yang, Xiang An, Kun Wu, Yongle Zhao, Weimo Deng, Zimin Ran,
  Yumeng Wang, Ziyong Feng, Roy Miles, Ismail Elezi, and Jiankang Deng.
\newblock Region-based cluster discrimination for visual representation
  learning.
\newblock In \emph{ICCV}, 2025.

\bibitem[Yin et~al.(2025)Yin, Si, and Wang]{yin2025clearsight}
Hao Yin, Guangzong Si, and Zilei Wang.
\newblock {ClearSight}: Visual signal enhancement for object hallucination
  mitigation in multimodal large language models.
\newblock In \emph{Proceedings of the IEEE/CVF Conference on Computer Vision
  and Pattern Recognition}, pages 14625--14634, 2025.

\bibitem[Zhang et~al.(2025{\natexlab{a}})Zhang, Wu, Li, Shang, Xia, Huang,
  Zhang, Dong, Zhang, Wang, Tan, and Wei]{zhang2025latentsketchpad}
Huanyu Zhang, Wenshan Wu, Chengzu Li, Ning Shang, Yan Xia, Yangyu Huang, Yifan
  Zhang, Li Dong, Zhang Zhang, Liang Wang, Tieniu Tan, and Furu Wei.
\newblock Latent sketchpad: Sketching visual thoughts to elicit multimodal
  reasoning in mllms.
\newblock \emph{arXiv:2510.24514}, 2025{\natexlab{a}}.

\bibitem[Zhang et~al.(2024{\natexlab{a}})Zhang, Singh, Liu, Liu, Yu, Gao, and
  Zhao]{zhang2024pasta}
Qingru Zhang, Chandan Singh, Liyuan Liu, Xiaodong Liu, Bin Yu, Jianfeng Gao,
  and Tuo Zhao.
\newblock Tell your model where to attend: Post-hoc attention steering for
  {LLMs}.
\newblock In \emph{International Conference on Learning Representations},
  2024{\natexlab{a}}.

\bibitem[Zhang et~al.(2025{\natexlab{b}})Zhang, Gao, Zhang, Li, Zhang, Liu,
  Yuan, Wu, Jia, Zhu, et~al.]{zhang2025adaptive}
Xintong Zhang, Zhi Gao, Bofei Zhang, Pengxiang Li, Xiaowen Zhang, Yang Liu, Tao
  Yuan, Yuwei Wu, Yunde Jia, Song-Chun Zhu, et~al.
\newblock Adaptive chain-of-focus reasoning via dynamic visual search and
  zooming for efficient vlms.
\newblock \emph{arXiv:2505.15436}, 2025{\natexlab{b}}.

\bibitem[Zhang et~al.(2026)Zhang, Lu, Yin, Fu, Chen, Hu, Wen, Jiang, Liu,
  Zhang, et~al.]{zhang2025thyme}
Yi-Fan Zhang, Xingyu Lu, Shukang Yin, Chaoyou Fu, Wei Chen, Xiao Hu, Bin Wen,
  Kaiyu Jiang, Changyi Liu, Tianke Zhang, et~al.
\newblock Thyme: Think beyond images.
\newblock In \emph{ICLR}, 2026.

\bibitem[Zhang et~al.(2024{\natexlab{b}})Zhang, Zhang, Li, Zhao, Karypis, and
  Smola]{zhang2024multimodal}
Zhuosheng Zhang, Aston Zhang, Mu Li, Hai Zhao, George Karypis, and Alex Smola.
\newblock Multimodal chain-of-thought reasoning in language models.
\newblock \emph{Transactions on Machine Learning Research}, 2024{\natexlab{b}}.

\bibitem[Zheng et~al.(2024)Zheng, Yin, Xie, Sun, Huang, Yu, Cao, Kozyrakis,
  Stoica, Gonzalez, Barrett, and Sheng]{zheng2024sglang}
Lianmin Zheng, Liangsheng Yin, Zhiqiang Xie, Chuyue Sun, Jeff Huang, Cody~Hao
  Yu, Shiyi Cao, Christos Kozyrakis, Ion Stoica, Joseph~E. Gonzalez, Clark
  Barrett, and Ying Sheng.
\newblock {SGLang}: Efficient execution of structured language model programs.
\newblock In \emph{NeurIPS}, 2024.

\bibitem[Zheng et~al.(2026)Zheng, Yang, Hong, Zhao, Xu, Yang, Shen, and
  Yu]{zheng2025deepeyes}
Ziwei Zheng, Michael Yang, Jack Hong, Chenxiao Zhao, Guohai Xu, Le Yang, Chao
  Shen, and Xing Yu.
\newblock {DeepEyes}: Incentivizing "thinking with images" via reinforcement
  learning.
\newblock In \emph{ICLR}, 2026.

\bibitem[Zhu et~al.(2025)Zhu, Wang, Chen, Liu, Ye, Gu, Tian, Duan, Su, Shao,
  et~al.]{zhu2025internvl3}
Jinguo Zhu, Weiyun Wang, Zhe Chen, Zhaoyang Liu, Shenglong Ye, Lixin Gu, Hao
  Tian, Yuchen Duan, Weijie Su, Jie Shao, et~al.
\newblock {InternVL3}: Exploring advanced training and test-time recipes for
  open-source multimodal models.
\newblock \emph{arXiv:2504.10479}, 2025.

\end{thebibliography}
}

\clearpage
\appendix
\section*{\centering\LARGE Appendix}
\section{LoRe Inference Details}
\label{app:inference_details}

This supplementary material provides implementation and evaluation details that support the main paper. We begin with the inference-time LoRe processor, which preserves the standard VLM input and performs visual focusing inside the same generation request. Algorithm~\ref{alg:supmat_lore_procedure} summarizes the runtime procedure from multimodal prefilling to final answer extraction.

\begin{algorithm}[!htbp]
    \caption{\textbf{LoRe single-turn inference algorithm.} LoRe converts a standard VLM request into region-guided decoding without adding a cropped image, extra visual tokens, or a second VLM call.}
    \label{alg:supmat_lore_procedure}
    \footnotesize
    \begin{algorithmic}[1]
        \REQUIRE Image $I$, question $Q$, VLM $\pi_\theta$, boost $\gamma$, interval $K$, window $W$
        \ENSURE Response with final answer in \texttt{<ans>}...\texttt{</ans>}
        \STATE Prefill $(I,Q)$ once and cache visual/text tokens as $C_\ell^{(0)}$.
        \STATE Record visual-token positions in the multimodal prefix.
        \STATE Decode autoregressively from $C_\ell^{(0)}$.
        \STATE Every $K$ tokens, search the latest $W$ tokens for \texttt{<box>}...\texttt{</box>}.
        \IF{no valid region span is available}
            \STATE Continue standard VLM decoding without changing the cache.
        \ELSE
            \STATE Parse $R_q=(x_1,y_1,x_2,y_2)$ and project it to $\mathcal{P}(R_q)$.
            \STATE If $\mathcal{P}(R_q)$ is non-empty, scale its value-cache entries once by $\gamma$.
            \STATE Keep keys, token order, and attention masks unchanged.
            \STATE Resume the same generation request from the modified cache.
        \ENDIF
        \STATE Extract the final answer using the base-model evaluator.
    \end{algorithmic}
\end{algorithm}

\textbf{Prompt format.}
The evaluation pipeline uses a single user message that contains the image followed by a text prompt. The text prompt starts with the benchmark question and then appends the same LoRe-specific output constraints across V*, HR-Bench, and COCO-ObjQA. In the notation used throughout the paper, the prompt has the following structure:
\begin{center}
\small
\setlength{\tabcolsep}{6pt}
\renewcommand{\arraystretch}{1.08}
\begin{tabular}{p{0.92\linewidth}}
    \toprule
    \textbf{Prompt template} \\
    \midrule
    \begin{minipage}{0.9\linewidth}
    \textbf{\{question\}}\\[0.25em]
    \textbf{Instructions:}\\
    1. Locate the most relevant object or image region that helps answer this question.\\
    2. Output its bounding box as \texttt{<box>}x1, y1, x2, y2\texttt{</box>}.\\
    3. Explain why this region is relevant.\\
    4. Give the final answer as \texttt{<ans>}X\texttt{</ans>}, where $X$ is the predicted option.
    \end{minipage} \\
    \bottomrule
\end{tabular}
\end{center}
For Qwen2.5-VL, the box coordinates are interpreted as pixel coordinates with $(x_1,y_1)$ and $(x_2,y_2)$ denoting the top-left and bottom-right corners. For Qwen3-VL, the same prompt uses the model's relative coordinate convention in $[0,1000]$, where zero and 1000 correspond to the image boundary along each axis. This difference affects only the coordinate parser before token-grid projection. The single-turn message structure, region span, and answer span are otherwise unchanged.

\textbf{Region parsing and fallback.}
LoRe uses the same image-question prompt as the base VLM, with an additional instruction to emit a region span before the final answer in \texttt{<ans>}...\texttt{</ans>} format. During generation, the processor scans the generated token stream for the first complete \texttt{<box>}x1, y1, x2, y2\texttt{</box>} pattern. To keep this step inexpensive, it checks the stream every four generated tokens and searches only the most recent 128 generated tokens. If no valid span is found, if the span is malformed, or if the parsed box covers no visual token, LoRe leaves the cache unchanged and continues with standard VLM decoding. This fallback path is used consistently during both training rollouts and evaluation.

\textbf{Coordinate and token mapping.}
All cache updates are applied to visual-token positions, so the parsed region must first be aligned with the visual-token grid. The processor converts the parsed box to image pixel coordinates when the prompt uses a relative coordinate convention, clamps the box to the image boundary, and maps the covered area to the grid described in Section~\ref{sec:implicit}. For Qwen-style VLMs, LoRe reads the image grid metadata returned by the model processor. If the vision backbone applies spatial token merging, the logical grid is adjusted to match the number of visual tokens stored in the language-model prefix. The selected grid cells are then mapped to sequence positions in the prompt. When explicit visual-token positions are available from the prompt token IDs, LoRe uses them directly. Otherwise, it falls back to the contiguous image-token span inferred from the expected visual-token count.

\textbf{Cache modulation.}
The default implementation uses \texttt{cache\_values} mode. After the target visual-token sequence positions are identified, LoRe scales the corresponding value-cache entries by $\gamma$ and leaves the key cache unchanged. We use $\gamma=2.0$ by default and apply the update from the first decoder layer through the final decoder layer. The cache entries are mutated once, after the region is detected, so subsequent decoding can continue with the model's native attention kernels. We also implemented two diagnostic variants: an eager additive attention-bias mode and a \texttt{cache\_kv} mode that scales both keys and values. The main paper reports value-cache scaling because it provides a favorable accuracy-efficiency tradeoff while avoiding changes to attention-score computation.

\section{Training-free Parser Diagnostics}
\label{app:parser_diagnostics}

Table~\ref{tab:supmat_parser_diagnostics} reports a small diagnostic run for the training-free LoRe pipeline before RL. The purpose is to verify that the base VLM can usually expose a parsable region span and that the fallback path is rare under the same prompt format used by LoRe. This diagnostic is not used to compute the main benchmark accuracy. \textbf{Samples} gives the number of evaluated examples. \textbf{Parsed box} is the fraction of generations that contain a syntactically valid \texttt{<box>}x1, y1, x2, y2\texttt{</box>} span. \textbf{Cache update} is the fraction of examples for which the parsed box can be mapped to at least one visual token and therefore triggers value-cache modulation. \textbf{Fallback} is the complementary fraction in which LoRe leaves the cache unchanged because no valid update region is available. \textbf{Avg. selected} reports the average number of visual tokens selected by the parsed region among updated examples, while \textbf{Avg. visual} reports the average total number of visual tokens in the original multimodal prefix.

\begin{table}[!htbp]
    \centering
    \caption{\textbf{Parser and cache-update diagnostics before RL.} We run Qwen2.5-VL-7B on a 200-sample COCO-ObjQA diagnostic subset with max generation length 512, and $\gamma=2.0$.}
    \vspace{0.2cm}
    \label{tab:supmat_parser_diagnostics}
    \small
    \setlength{\tabcolsep}{4.5pt}
    \renewcommand{\arraystretch}{1.14}
    \definecolor{supDiagRule}{RGB}{190,178,166}
    \definecolor{supDiagHL}{RGB}{255,243,235}
    \begingroup
    \arrayrulecolor{supDiagRule}
    \resizebox{\linewidth}{!}{%
    \begin{tabular}{lcccccc}
        \toprule
        \textbf{Model} & \textbf{Samples} & \textbf{Parsed box} & \textbf{Cache update} & \textbf{Fallback} & \textbf{Avg. selected} & \textbf{Avg. visual} \\
        \midrule
        \rowcolor{supDiagHL}
        Qwen2.5-VL-7B & 200 & 97.5\% & 97.5\% & 2.5\% & 64.5 & 331.7 \\
        \bottomrule
    \end{tabular}
    }
    \arrayrulecolor{black}
    \endgroup
    \vspace{-0.2cm}
\end{table}

\section{Reinforcement Learning Details}
\label{app:rl_details}

\textbf{Reward components.}
For each rollout, the reward system first computes the answer reward $r_{\mathrm{acc}}$ using the task's answer type. Multiple-choice tasks are evaluated by extracting the final option from \texttt{<ans>}...\texttt{</ans>}, while numerical, OCR, Boolean, and string-style tasks use their corresponding matching rules. The auxiliary format reward $r_{\mathrm{fmt}}$ is binary: it is one only when the completion contains exactly one non-empty \texttt{<ans>}...\texttt{</ans>} span and any optional reasoning delimiters are balanced. This term supports answer extraction, while box presence and parseability are rewarded by $r_{\mathrm{loc}}$.

The localization reward $r_{\mathrm{loc}}$ is used as weak shaping rather than as the dominant objective. For visually grounded problem types, including grounding, spatial reasoning, and embodied tasks, it adds a small bonus when a parsable box is present. For bbox-supervised samples, it additionally rewards parseability and center proximity to the target box. Given a predicted box $b$ and a target box $b^\star$, the center component is
\[
    r_{\mathrm{center}}(b,b^\star)
    =
    \max\left(0,\;1-\frac{\|c(b)-c(b^\star)\|_2}{\sqrt{2}\,s}\right),
\]
where $c(\cdot)$ is the box center and $s$ is the largest absolute coordinate value, lower bounded by one. The localization term used in the experiments is
\[
    r_{\mathrm{loc}}
    =
    0.10\,r_{\mathrm{box}}
    +0.05\,r_{\mathrm{boxfmt}}
    +0.15\,r_{\mathrm{center}},
\]
where unavailable terms are set to zero. Here $r_{\mathrm{box}}$ is the box-presence reward for visually grounded samples, $r_{\mathrm{boxfmt}}$ is the parseability reward for bbox-supervised samples, and $r_{\mathrm{center}}$ is the center-proximity reward above. The final reward follows the residual form in the main paper:
\[
    R=\min\{1,\;r_{\mathrm{acc}}+(1-r_{\mathrm{acc}})(0.03\,r_{\mathrm{fmt}}+r_{\mathrm{loc}})\}.
\]
This form keeps answer correctness as the primary learning signal: auxiliary terms can improve partially correct trajectories, but they cannot push a correct answer above one or dominate $r_{\mathrm{acc}}$.

\textbf{GRPO hyperparameters.}
For each prompt, we sample $G=16$ trajectories with temperature 1.0 and maximum response length 4096. Rewards are transformed by $S_i=\mathrm{clip}(5(R_i-0.5),-c,c)$ before group normalization, matching the scaling and bias used in our training configuration. We use asymmetric PPO clipping with lower and upper bounds 0.2 and 0.28, respectively, and a KL coefficient of 0.01 against the frozen reference policy. The actor uses Adam with learning rate $1\times10^{-6}$, weight decay 0.01, $\beta_1=0.9$, $\beta_2=0.999$, gradient clipping at 1.0, and bfloat16 precision. Training uses 8 H200 GPUs: four GPUs serve SGLang rollouts, and four GPUs compute actor and reference-policy log probabilities with FSDP. Groups whose rewards are identical are filtered because they provide no relative advantage signal.

\section{COCO-ObjQA Construction Details}
\label{app:coco_details}

COCO-ObjQA is derived from MS COCO instance annotations. For each image, we enumerate annotated object instances and keep target categories that appear exactly once in the image. This filtering reduces ambiguity: when a generated question refers to the target category, there is a single annotated instance that can serve as the intended visual evidence. For each retained target, we crop the annotated object from the image and provide both the full image and the crop to Gemini 3, together with the category name and the ground-truth box. The prompt asks Gemini 3 to generate a concise multiple-choice question about visible attributes of the focused object, without referring to the crop, bounding box, or object location.

\begin{figure}[t]
    \centering
    \includegraphics[width=1\linewidth]{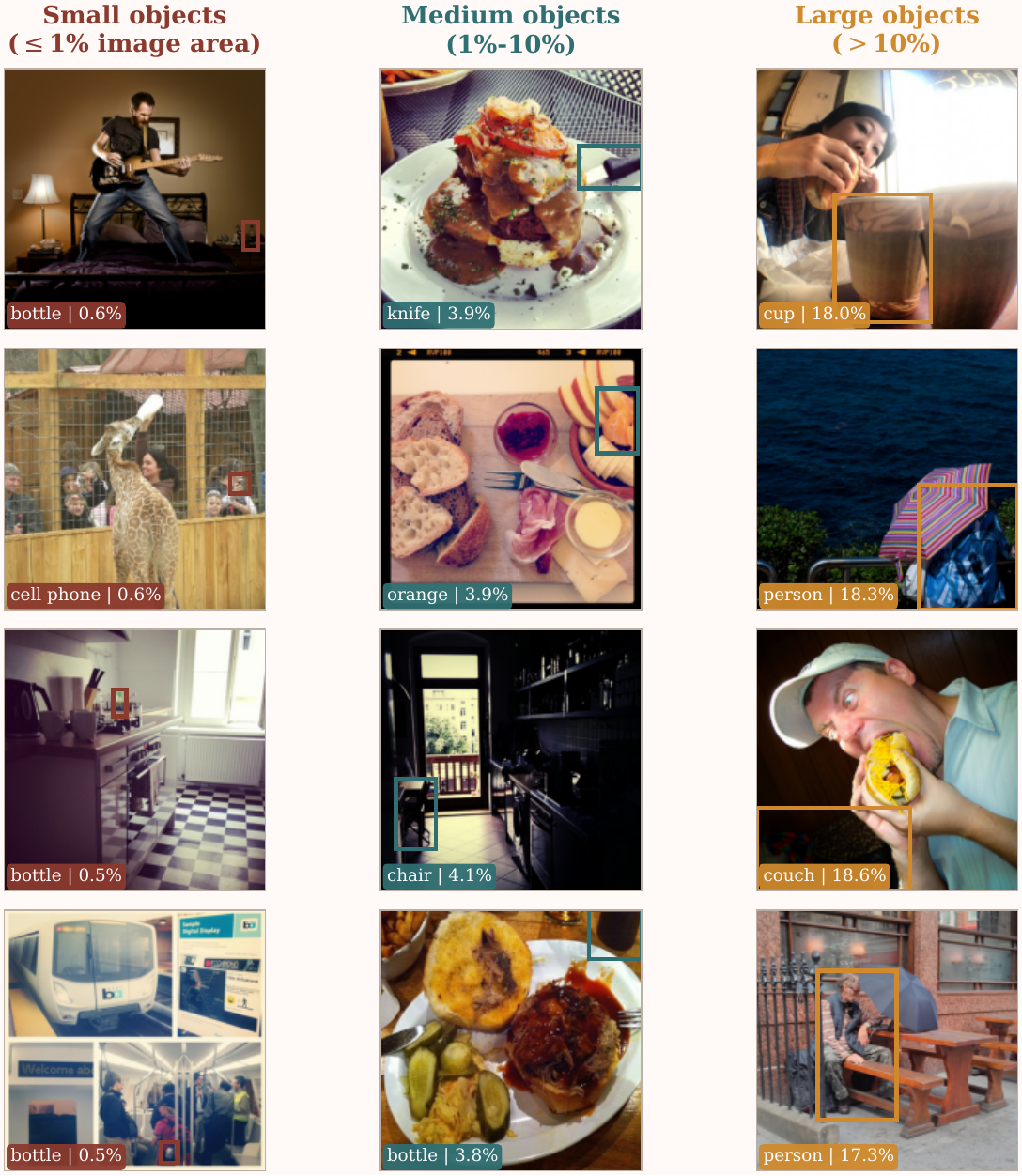}
    \vspace{-0.2cm}
    \caption{\textbf{COCO-ObjQA examples across target-object scales.} We visualize representative full-image samples from the small, medium, and large splits. Boxes mark the target COCO instances used to generate object-centric questions, and labels show the category and target-area ratio. The boxes are used for dataset construction and localization diagnostics, while models are evaluated on the original full image.}
    \label{fig:supmat_coco_objqa_dataset_visualization}
    \vspace{-0.2cm}
\end{figure}

We first check that the generated text follows the expected six-line structure: one question line, four answer-option lines, and one answer line. An independent GPT-5.4 check then assesses whether the question is answerable from the original full image and whether the labeled option is correct and unambiguous. Flagged problems are removed, and the remaining samples are ranked by estimated answerability or labeling risk for targeted manual inspection. Confirmed failures are removed before the final benchmark is formed. This inspection targets high-risk samples rather than a random subset and does not establish an unbiased benchmark-wide error estimate. Generated questions may still inherit model biases.

Each benchmark record stores the full-image path, question, options, answer label, COCO box, category, and target-area ratio. Standard evaluation supplies only the original full image and multiple-choice question. Crops and boxes support construction and localization diagnostics. Only the explicitly labeled oracle-box ablation uses reference boxes to guide decoding.

The targeted manual audit covers the 319 highest-risk samples after automatic filtering. Before removal of confirmed failures, 94.7\% of these questions are answerable and 90.6\% have the correct ground-truth option. These rates describe the selected high-risk subset, not a random estimate of label quality across COCO-ObjQA. The final benchmark statistics below are computed after automatic and manual filtering.

The size split is computed from the target-area ratio, defined as the COCO box area divided by the full image area. Following the main paper, small, medium, and large objects correspond to target-area ratios of $\leq$1\%, 1--10\%, and $>$10\%, respectively. COCO-ObjQA contains 8,537 QA pairs from 4,107 images across all 80 COCO categories, with 1,743 small-object examples, 3,036 medium-object examples, and 3,758 large-object examples. This stratification is used only for evaluation, and the model receives the same full-image input format across all size groups.

\section{Evaluation and Latency Protocol}
\label{app:evaluation_details}

\textbf{Answer extraction.}
The three primary benchmarks use exact-match accuracy on the final option. The evaluator first extracts the option from the model's \texttt{<ans>}...\texttt{</ans>} span. If a strict answer span is unavailable, it falls back to conservative option matching near the end of the response. The same extraction logic is applied to base VLMs and LoRe variants.

\textbf{LoRe evaluation settings.}
Unless otherwise stated, LoRe evaluation uses value-cache scaling with $\gamma=2.0$, all decoder layers enabled, a bbox check interval of four generated tokens, and a decode window of 128 generated tokens for region detection. Evaluation generation uses the same image resolution and prompt formatting as the corresponding base model, except for the additional instruction to emit the region span. The maximum generation length for benchmark evaluation is 2048 tokens.

\textbf{Latency measurement.}
Latency is reported as average per-sample wall-clock generation time. All methods in Table~\ref{tab:sota} are run on the same H200 GPU machine with batch size one, and without an external inference engine such as SGLang or vLLM. For LoRe and the corresponding direct VLM baselines, timing starts immediately before model generation and stops after output tokens are produced. Dataset loading, metric aggregation, and JSON serialization are excluded. The LoRe timing includes box parsing and cache modulation, so the reported latency reflects the additional inference work introduced by the method. For explicit \emph{thinking with images} methods, latency includes the full multi-step reasoning path, including additional VLM interactions and visual operations, when those operations are part of the method's inference procedure.

\textbf{Localization diagnostics.}
For standard COCO-ObjQA evaluation, the ground-truth box is used only after generation to measure localization quality. COCO-ObjQA$_{\mathrm{IoU}}$ in Table~\ref{tab:ablation_training} averages IoU over all evaluated examples, assigning zero when no valid box is produced. It is not used to choose the final answer. The separate oracle-box ablation substitutes COCO boxes on COCO-ObjQA and SAM2-derived~\cite{ravi2025sam2} pseudo-ground-truth regions on V* and HR-Bench while retaining the same RL checkpoint, $\gamma=2$, and decoding procedure. These reference regions provide a localization diagnostic rather than a strict performance bound.

\section{Additional Experiments and Diagnostics}
\label{app:additional_experiments}

\subsection{Controlled RL and crop-based comparisons}
\label{app:controlled_comparisons}

Table~\ref{tab:ablation_controls} distinguishes training without modulation from disabling modulation in an already trained LoRe policy. For $\gamma_{\mathrm{train}}=\gamma_{\mathrm{test}}=1$, GRPO still learns to emit a box and reasoning trace, but neither training nor evaluation modifies the cache. For the two $\gamma_{\mathrm{train}}=2$ rows, the checkpoint, prompt, box-generation process, and decoding procedure are identical. Only the test-time value scaling changes. Disabling scaling therefore evaluates the trained policy without the cache intervention it encountered during learning. This is distinct from training a policy without modulation from the outset.

The crop-based GRPO control uses the same Qwen2.5-VL-7B initialization, training data, recipe, and objectives as LoRe. It emits one box, receives the selected crop as a re-encoded second visual observation, and then answers. The comparison changes how the selected region is returned to the model. Crop re-encoding performs better on COCO-ObjQA, while LoRe performs better on V* and HR-Bench without adding a visual observation. These results support an accuracy--computation tradeoff rather than uniform dominance over explicit visual processing.

\subsection{Cache-modulation variants}
\label{app:modulation_variants}

We compare three interventions on the same Qwen2.5-VL-7B base model before RL (Table~\ref{tab:modulation_variants}). Additive score bias adds $\log\gamma$ to the selected visual-token logits, multiplying their attention odds by $\gamma$. The implemented eager attention path explicitly computes scores and softmax, preventing the use of Flash Attention on modified layers. Value-only scaling and joint key--value scaling instead mutate the selected cache entries once and retain the native optimized kernels. Additive bias and value-only scaling use $\gamma=2$. Joint key--value scaling uses $\gamma=1.01$ because stronger scaling was unstable, so this is a comparison of the evaluated configurations rather than a matched-strength sweep.

\begin{table}[!htbp]
    \centering
    \caption{\textbf{Training-free modulation variants.} Accuracy is reported in percent and latency in seconds. The final columns give direct changes in selected-token attention mass (percentage points) and attention entropy (nats).}
    \label{tab:modulation_variants}
    \small
    \setlength{\tabcolsep}{4pt}
    \renewcommand{\arraystretch}{1.12}
    \resizebox{\linewidth}{!}{%
    \begin{tabular}{@{}lccccccc@{}}
        \toprule
        \textbf{Intervention} & $\gamma$ & \textbf{V*} & \textbf{HR-Bench} & \textbf{COCO-ObjQA} & \textbf{Latency} & $\Delta$ \textbf{mass} & $\Delta$ \textbf{entropy} \\
        \midrule
        Additive score bias & 2 & 76.8 & 68.2 & 89.3 & 2.71 & +0.24 & +0.007 \\
        Value scaling & 2 & \textbf{77.4} & \textbf{70.2} & \textbf{89.4} & \textbf{2.16} & 0.00 & 0.000 \\
        Key--value scaling & 1.01 & 74.3 & 68.5 & 89.3 & 2.33 & +0.26 & +0.005 \\
        \bottomrule
    \end{tabular}%
    }
\end{table}

Value-only scaling gives the highest accuracy and lowest latency among these configurations. Its zero direct change in attention mass and entropy follows from leaving queries and keys unchanged at the intervention: the attention output changes through the values, not the current attention probabilities. Subsequent attention can still change indirectly through the modified residual state. In contrast, scaling a key multiplies its logit by $\gamma$, increasing positive logits while making negative ones more negative, which explains why it is not equivalent to a uniformly positive attention bias.

\subsection{Fallback behavior before and after RL}
\label{app:fallback_results}

Table~\ref{tab:fallback_results} complements the 200-sample, training-free parser diagnostic in Appendix~\ref{app:parser_diagnostics} with benchmark-level results before and after RL. A fallback occurs when the generated box is missing, malformed, or cannot be mapped to visual tokens. The cache is left unchanged and the same policy continues decoding. These samples remain in the overall accuracy. A valid but inaccurate box is not a fallback and may still amplify irrelevant evidence.

\begin{table}[!htbp]
    \centering
    \caption{\textbf{Fallback frequency and answer accuracy for Qwen2.5-VL-7B.} All entries are percentages. Fallback and non-fallback accuracy are measured within their respective subsets.}
    \label{tab:fallback_results}
    \small
    \setlength{\tabcolsep}{5pt}
    \renewcommand{\arraystretch}{1.12}
    \begin{tabular}{@{}llcccc@{}}
        \toprule
        \textbf{Policy} & \textbf{Benchmark} & \textbf{Fallback rate} & \textbf{Fallback acc.} & \textbf{Non-fallback acc.} & \textbf{Overall acc.} \\
        \midrule
        Pre-RL & V* & 11.0 & 61.9 & 79.3 & 77.4 \\
        Pre-RL & HR-Bench & 8.5 & 64.7 & 70.7 & 70.2 \\
        Pre-RL & COCO-ObjQA & 2.5 & 79.8 & 89.6 & 89.4 \\
        \midrule
        RL-trained & V* & 6.8 & 69.2 & 83.6 & 82.6 \\
        RL-trained & HR-Bench & 5.0 & 70.0 & 73.8 & 73.6 \\
        RL-trained & COCO-ObjQA & 1.5 & 66.4 & 92.0 & 91.6 \\
        \bottomrule
    \end{tabular}
\end{table}

RL reduces the fallback rate on all three benchmarks, although fallback cases retain lower accuracy than non-fallback cases. In these measurements, every syntactically valid box was successfully mapped, including on HR-Bench's 4K and 8K splits. The mapping reads each image's dynamic grid metadata and spatial merge factor, so varying resolution and aspect ratio do not require a fixed patch grid. Patch quantization can still limit localization precision for very small regions.

\subsection{Oracle localization and remaining errors}
\label{app:oracle_analysis}

We use the same RL-trained Qwen2.5-VL-7B checkpoint and decoding procedure for predicted-box and oracle-box evaluation. Reference regions are COCO annotations on COCO-ObjQA and SAM2-derived pseudo-ground-truth boxes on V* and HR-Bench, as described in Section~\ref{sec:ablation}. Table~\ref{tab:oracle_boost} varies the inference-time boost factor while holding the reference regions and checkpoint fixed. The best evaluated factor remains $\gamma=2$, indicating that stronger amplification alone does not remove the localization gap.

\begin{table}[!htbp]
    \centering
    \caption{\textbf{Boost strength with reference regions.} All rows use the same RL checkpoint trained at $\gamma=2$. Scores are average accuracy (\%).}
    \label{tab:oracle_boost}
    \small
    \setlength{\tabcolsep}{10pt}
    \renewcommand{\arraystretch}{1.12}
    \begin{tabular}{@{}lcccc@{}}
        \toprule
        \textbf{Region source} & $\gamma_{\mathrm{test}}$ & \textbf{V*} & \textbf{HR-Bench} & \textbf{COCO-ObjQA} \\
        \midrule
        Predicted box & 2 & 82.6 & 73.6 & 91.6 \\
        Oracle box & 1.5 & 83.5 & 75.0 & 92.5 \\
        Oracle box & 2 & \textbf{84.1} & \textbf{75.8} & \textbf{93.0} \\
        Oracle box & 2.5 & 83.8 & 75.4 & 92.7 \\
        Oracle box & 3 & 83.2 & 74.7 & 92.2 \\
        \bottomrule
    \end{tabular}
\end{table}

Table~\ref{tab:oracle_errors} compares correctness per example at $\gamma=2$. ``Predicted only'' denotes examples answered correctly only with the predicted box, while ``Oracle only'' denotes errors corrected by the reference box. Recovery is the fraction of predicted-box errors corrected by the oracle, computed as $\mathrm{Oracle\ only}/(\mathrm{Oracle\ only}+\mathrm{Both\ wrong})$.

\begin{table}[!htbp]
    \centering
    \caption{\textbf{Per-example oracle comparison.} The four outcome columns partition each benchmark and sum to 100\%. Recovery is conditioned on predicted-box errors.}
    \label{tab:oracle_errors}
    \small
    \setlength{\tabcolsep}{7pt}
    \renewcommand{\arraystretch}{1.12}
    \begin{tabular}{@{}lccccc@{}}
        \toprule
        \textbf{Benchmark} & \textbf{Both correct} & \textbf{Predicted only} & \textbf{Oracle only} & \textbf{Both wrong} & \textbf{Recovery} \\
        \midrule
        V* & 80.5 & 2.1 & 3.6 & 13.8 & 20.7 \\
        HR-Bench & 71.5 & 2.1 & 4.3 & 22.1 & 16.3 \\
        COCO-ObjQA & 89.8 & 1.8 & 3.2 & 5.2 & 38.1 \\
        \bottomrule
    \end{tabular}
\end{table}

Reference boxes correct more errors than they introduce, but some predictions regress and substantial errors remain with either region source. The comparison therefore identifies scope for improved localization without assigning all residual errors to it. Recovering missing visual detail or improving subsequent evidence integration may be necessary for further gains. For relative-position questions, a single box may omit one of the compared objects, select a visually similar distractor, or include irrelevant context between distant objects. The multi-region results in Figure~\ref{fig:ablation_kv_focus} show that independently adding boxes does not reliably solve this evidence-integration problem.

\subsection{Transfer beyond object-attribute questions}
\label{app:transfer_results}

We evaluate the training-free LoRe pipeline on MathVista-testmini~\cite{lu2024mathvista}, We-Math~\cite{qiao2025wemath}, and TIR-Bench~\cite{li2025tirbench}. Both rows in Table~\ref{tab:transfer_results} use the same Qwen2.5-VL-7B base checkpoint. LoRe applies its default value-cache update directly to the original full image, with no task-specific training, cropping, image re-encoding, or additional VLM call.

\begin{table}[!htbp]
    \centering
    \caption{\textbf{Training-free transfer to additional reasoning tasks.} Scores are accuracy (\%).}
    \label{tab:transfer_results}
    \small
    \setlength{\tabcolsep}{9pt}
    \renewcommand{\arraystretch}{1.12}
    \begin{tabular}{@{}lccc@{}}
        \toprule
        \textbf{Method} & \textbf{MathVista-testmini} & \textbf{We-Math} & \textbf{TIR-Bench} \\
        \midrule
        Qwen2.5-VL-7B & 68.6 & 61.3 & 16.0 \\
        LoRe (pre-RL) & 71.3 & 64.7 & 16.2 \\
        \bottomrule
    \end{tabular}
\end{table}

LoRe improves MathVista and We-Math by 2.7 and 3.4 percentage points, respectively, supporting the use of evidence focusing for diagrams, labels, and quantities. The gain on TIR-Bench is only 0.2 points. This limited improvement is consistent with the method's scope: amplifying a fixed visual representation cannot create the transformed observations required by operations such as rotation or flipping. These results support transfer across evidence-selection tasks, rather than generality across arbitrary image manipulations.

\section{Limitations}
\label{app:limitations}

LoRe assumes that the model can emit a useful region before completing the final answer. Missing, malformed, or unmappable regions leave the cache unchanged, and decoding continues with the same policy, whereas a valid but inaccurate box can amplify unhelpful evidence. The oracle-box comparison in Figure~\ref{fig:ablation_kv_focus} indicates scope for improved localization but does not explain every remaining error.

LoRe operates on visual tokens already present in the original prefix. It cannot recover discarded detail or implement transformations such as rotation and flipping that require a new observation. A single enclosing box may cover multiple objects, but cannot weight them independently and may include distractors between distant targets. Multiple independently predicted regions do not consistently improve the reported results. Selective crop or zoom use could complement implicit focusing, but would require a routing mechanism and additional image encoding outside the current single-turn design.

COCO-ObjQA diagnoses single-target evidence selection across object sizes, but its generated questions and targeted validation do not guarantee exhaustive label correctness or coverage of relational reasoning. The broader evaluation is limited to single-image question answering. Multi-image, video, and interactive settings would require additional handling of temporal and cross-image token positions.

\section{Broader Impacts}
\label{app:broader_impacts}

LoRe aims to improve fine-grained multimodal reasoning while reducing the extra inference cost introduced by explicit visual-tool use. This can make visual reasoning systems more practical and more compute-efficient, especially in settings where a model needs to answer questions about localized evidence without repeatedly cropping or re-encoding images. At the same time, stronger and cheaper multimodal reasoning can also inherit the broader risks of VLMs, including incorrect visual interpretations, over-reliance on model outputs, and potential misuse in automated decision-making. LoRe does not by itself provide safety guarantees, so deployment in sensitive domains should include task-specific validation, human oversight, and safeguards appropriate to the application.

\end{document}